\documentclass{ieeeaccess}
\pdfoutput=1
\usepackage{cite}
\usepackage{amsmath,amssymb,amsfonts}
\usepackage{graphicx}
\usepackage{textcomp}
\def\BibTeX{{\rm B\kern-.05em{\sc i\kern-.025em b}\kern-.08em
    T\kern-.1667em\lower.7ex\hbox{E}\kern-.125emX}}
\usepackage{algorithm}
\usepackage{algorithmic}
\newcommand{\ie}{{\it i.e.}}
\newcommand{\eg}{{\it e.g.}}
\newcommand{\omni}{360\textdegree}
\newcommand{\etal}{{\it et al.}}

\newcommand{\Eref}[1]{Eq.~(\ref{#1})}
\newcommand{\Fref}[1]{Fig.~\ref{#1}}
\newcommand{\Sref}[1]{Section~\ref{#1}}
\newcommand{\Cref}[1]{Chapter~\ref{#1}}
    
\begin{document}
\history{Date of publication xxxx 00, 0000, date of current version August 18, 2022.}
\doi{10.1109/ACCESS.2022.DOI}

\title{Saliency-based Multiple Region of Interest Detection from a Single \omni~ image}
\author{\uppercase{Yuuki Sawabe}\authorrefmark{1}  \IEEEmembership{Student Member, IEEE},
\uppercase{Satoshi Ikehata}\authorrefmark{2} 
\IEEEmembership{Member, IEEE}, \\
\uppercase{Kiyoharu Aizawa} \authorrefmark{3} 
\IEEEmembership{Fellow, IEEE}.
\address[1]{Graduate School of Information Science and Technology, The University of Tokyo (e-mail: sawabe@hal.t.u-tokyo.ac.jp)}
\address[2]{National Institute of Informatics (e-mail: sikehata@nii.ac.jp)}
\address[3]{Graduate School of Information Science and Technology, The University of Tokyo (e-mail: aizawa@hal.t.u-tokyo.ac.jp)}}

\corresp{Corresponding author: Yuuki Sawabe (e-mail: sawabe@hal.t.u-tokyo.ac.jp).}

\markboth
{Yuuki Sawabe \headeretal: Saliency-based Multiple Region of Interest Detection from a Single \omni~ image}
{Yuuki Sawabe \headeretal: Saliency-based Multiple Region of Interest Detection from a Single \omni~ image}

\begin{abstract}
\omni~ images are informative -- it contains omnidirectional visual information around the camera. 
However, the areas that cover a~\omni~image is much larger than the human's field of view, therefore important information in different view directions is easily overlooked. To tackle this issue, we propose a method for predicting the optimal set of Region of Interest (RoI) from a single \omni~image using the visual saliency as a clue. 
To deal with the scarce, strongly biased training data of existing single \omni~image saliency prediction dataset, we also propose a data augmentation method based on the spherical random data rotation. From the predicted saliency map and redundant candidate regions, we obtain the optimal set of RoIs considering both the saliency within a region and the Interaction-Over-Union (IoU) between regions. We conduct the subjective evaluation to show that the proposed method can select regions that properly summarize the input \omni~image.
\end{abstract}


\titlepgskip=-15pt

\maketitle

\section{Introduction}
\begin{figure}[t]
    \centering
    \includegraphics[width=\linewidth]{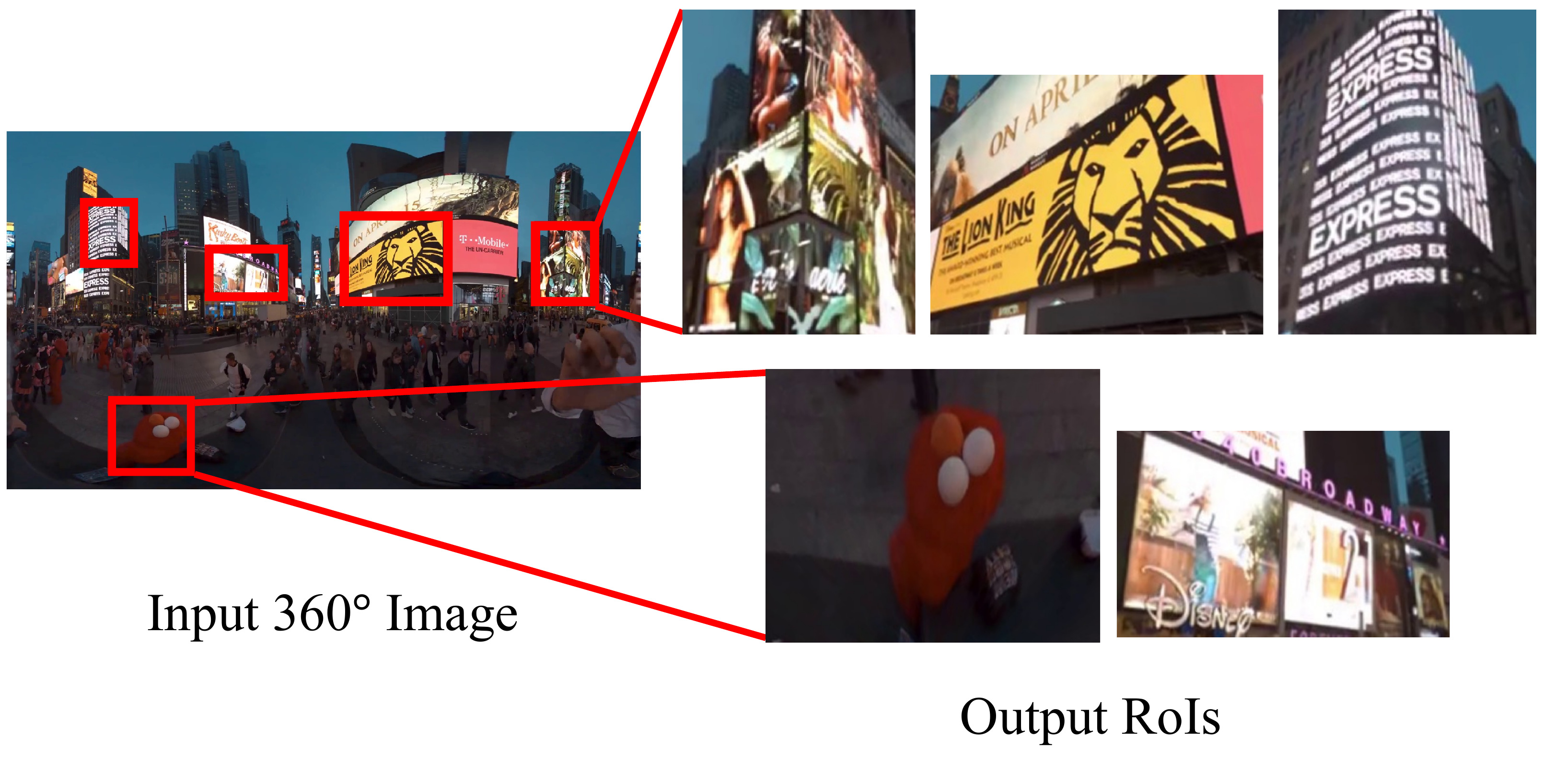}
    \vspace{-6mm}
    \caption{The task proposed in this study. Detecting the Region of Interest on the right from the \omni~ ERP image on the left.}
    \vspace{-6mm}
    \label{teaser}
\end{figure}
\begin{figure*}[t]
  \begin{center}
   \includegraphics[width=\linewidth]{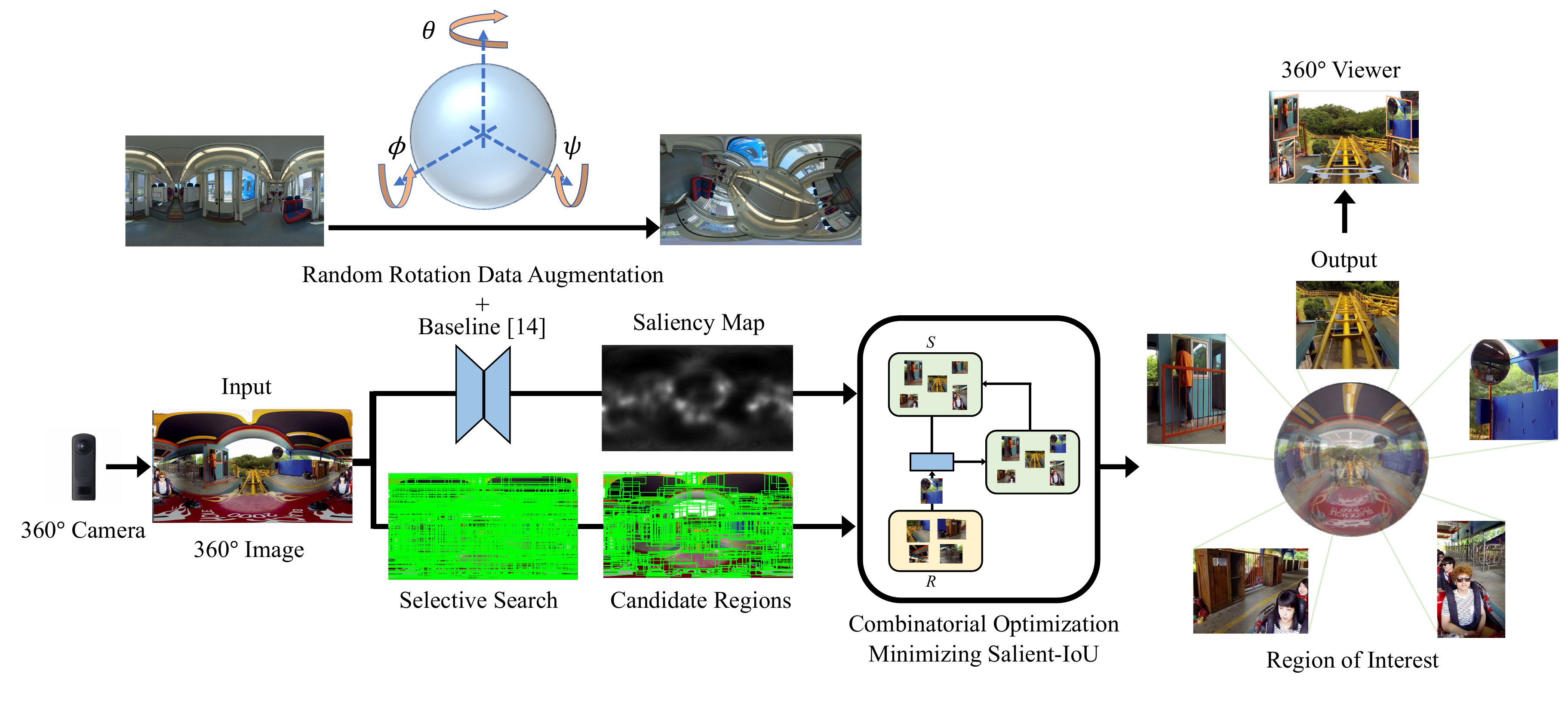}
   
  \end{center}
  \vspace{-4mm}
  \caption{A framework for detecting multiple Region of Interest from a single\omni~ image}
  \vspace{-4mm}
  \label{teian_all}
\end{figure*}

In recent years, \omni~images and videos have attracted significant attention owing to their advantage in terms of their omnidirectional information over the perspective images whose typical field of view~(FoV) is less than 65\textdegree~(\ie, normal field-of-view (NFoV)~\cite{pano2vid}). 
However, the human vision system does not have such a large FoV and achieving an efficient way of viewing the \omni~ image is an important problem in preventing important information in an \omni~image from being overlooked. 

To tackle this problem, one effective solution is to summarize the entire \omni~image into a set of Region of Interest (RoI) and present them to the viewer when browsing \omni~images or videos~\cite{pano2vid,snap_angle}. In our context, RoI is defined as the local region of a small FoV that might draw the attention of the viewer, and is expected to represent the important information in the input \omni~image or video. In the early work by Su~\etal~\cite{pano2vid}, which first attempted to extract an RoI from a \omni~ video, it was assumed there was only a single RoI in each frame, which can be problematic when multiple important items are observed at the same time. In addition, the input must have multiple frames from a \omni~video because it utilizes temporal information to decide which view is good to crop. To tackle this issue, Xiong~\etal~\cite{snap_angle} have recently proposed a method for extracting multiple RoIs from a single \omni~ image by projecting the \omni~image onto a unit cube and optimizing the rotation of the cube such that the extracted RoIs are placed at the centers of its faces. 
Although it allows multiple RoIs from a single image, the FoV covered by each face of the cube (\ie, 90\textdegree~ ) is larger than human's FoV and the number (\ie, six) and relative positions of the RoIs (\ie, two different faces of cubes) are always fixed. Therefore, it is highly likely that the extracted RoIs will be either of redundant or scarce. 

To properly support a more effective \omni~ image browsing, we propose a method that takes a single \omni~ image as input and outputs the fixed number of RoIs {\it whose positions, and corresponding FoVs are adaptively selected}. To the best of our knowledge, this is the first attempt to predict multiple NFoV RoIs from a single 360-degree image without constraining their sizes and positions.

Example results are illustrated in~\Fref{teaser}. In this example, we extracted five RoIs of varying size from the input \omni~image in ERP (EquiRectangular Projection) format. Our method firstly divides the input ERP image into multiple overlapping candidate rectangular regions through Selective Search~\cite{selective_search} and find the optimal fixed-sized (\ie, five in this example) subset by optimizing our new evaluation function, namely, {\it Salient-IoU}, which considers the saliency values within each region and the IoU which measures the overlap between two regions. Saliency~\cite{saliency_neuro} is a quantitative measure of attracting human visual attention and has been used in image recognition, object detection, robotics, and advertising design. In our framework, the accurate prediction of saliency values is a critical component. The prediction of saliency originated from Itti~\etal's method~\cite{itti}, which was the first to introduce a bottom-up model of human visual attention, and in recent years, deep neural network models based on gaze information have become mainstream~\cite{deepgaze2}. However, existing saliency map (a 2-D image that contains the saliency value at each pixel) prediction is mainly for perspective projection cameras, and the performance for \omni~ images is fairly limited~\cite{salgan-360,baseline,salnet360,salgcn}. To overcome this limitation, we propose a simple but effective new data augmentation method using a random rotation on a unit sphere which deals with the scarce, strongly biased training data of existing single \omni~image saliency prediction dataset~\cite{salient360,salient360-image}. 

Owing to a lack of existing research on the multiple RoI extraction from a single \omni~image for the summarization purpose, we evaluate our method based on the user study on Amazon Mechanical Turk\cite{amt} and show that our results match the human's intuition.

A summary of our contributions is as follows:
\begin{itemize}
    \item To the best of our knowledge, this is the first work that predicts multiple RoIs from a single 360-degree image without constraining the size and position of the RoIs. 
    \item We present the simple but effective data augmentation strategy for improving single \omni~image saliency prediction by which our saliency prediction significantly outperforms the state of the art.
    \item We conducted a user evaluation on the Amazon Mechanical Turk to illustrate the performance of our RoI prediction.
\end{itemize}
\section{Related works}
 \subsection{\omni~information summarization}
It is an important task to efficiently present \omni~images/videos to the limited human's view. However, as already mentioned, there is no prior work which predicted multiple RoIs from a single \omni~image, to the best of knowledge. On the other hand, the summarization of \omni~video had been an active topic that is to present the most important view direction at each frame. For instance, Su~\etal~\cite{pano2vid} proposed an optimization-based algorithm to find a path over the spatio-temporal glimpses that maximize the accumulated capture-worthiness score while obeying a smooth camera motion constraint. This work was later extended to allow more general camera control such as zooming~\cite{Su2017}. Benefiting from deep-learning-based object detection methods, Deep 360 Pilot~\cite{deep_360_pilot} presented an object-centric deep-learning-based agent for piloting through 360° sports videos automatically. While typical \omni~video summarizing task targets to find the optimal spatial camera trajectories, Lee~\etal~\cite{Lee2018} also addressed story-based temporal summarization by leveraging the memory networks. 
 
 Recently, Wang~\etal~\cite{Wang2020} have presented {\it Transitioning360}, a tool for 360° video navigation on 2-D displays by transitioning between multiple NFoV views that track potentially interesting targets or events. They combined saliency map, optical flow and object instances computed from the input \omni~video and optimized the virtual NFoV paths based on both contents and temporal smoothness. While this work also uses the saliency information to detect important regions in \omni~video and presented viewers multiple RoIs at the same time, the predicted RoIs are basically object centric, which relies on the specific object categories and more importantly this method cannot be applied to a single frame. On the other hand, our method introduces Selective Search~\cite{selective_search} to find perceptually important, non-object-centric candidate regions and is completely applicable to a single \omni~image.
 
\subsection{Saliency map prediction from \omni~ image} \label{sec:relatedwork_sal}
In contrast to the saliency map prediction on normal perspective images, estimating the saliency map of a \omni~image in ERP format is more challenging due to distortions caused by projection from a sphere to a plane. 

Traditionally, the existing bottom-up model of human visual attention~\cite{itti} was extended to \omni~images in ERP format and the saliency map was predicted based on that model~\cite{Bur2006}. However the method that only considers low level visual features limits its prediction accuracy. 

The data-driven, image-based saliency detection was made possible by the advent of public data sets. The first moderate-scale public datasets of \omni~images with associated eye and head movement data had been presented by Rai~\etal~\cite{Rai2017} which consists of sixty different \omni~images and gaze information by at least 40 observers. This dataset was later followed up by Erwan~\etal~\cite{Erwan2018} to extend it to \omni~videos. Using these dataset, Monroy~\etal~\cite{salnet360} presented the first data-driven saliency map prediction method which takes a \omni~image as input and splits it into six patches to be fed to convolutional neural networks (CNN). Chen~\etal~\cite{Cheng2018} presented a spatio-temporal nework to predict \omni~video saliency with cube-padding technique to avoid the sphere-to-plane distortion problem. Zhang~\etal~\cite{Zhang2018} presented the \omni~ video saliency detection by a spherical convolution neural network trained on 104 \omni~videos viewed by 27 human subjects. Chao~\etal~\cite{salgan-360} extended the perspective saliency prediction model trained with adversarial examples~\cite{salgan} to \omni~images and won Salient360! Grand Challenges at ICME'18 in the task of prediction of head and eye saliency, and this method became a touchstone in the domain of the \omni~ image saliency map prediction.

There are two most recent works leveraging state-of-the-art deep learning techniques. Haoran~\etal~\cite{salgcn} presented the \omni~saliency detection algorithm based on the graph convolutional neural networks. Specifically, a spherical graph signal was constructed from a ERP image and saliency map was generated from the spherical features on the graph. Martin~\etal~\cite{baseline} leveraged \omni-aware convolutions that represent kernels as patches tangent to the sphere where the panorama is projected, and a spherical loss function that penalizes prediction errors for each pixel depending on its coordinates in a gnomonic projection.

One of the remaining challenges is the limited amount of data compared to perspective images. In particular, data diversity is an important issue, and is also a cause of strong center bias, where the salient region is concentrated on the equator. However, no clear solution to this problem has been proposed yet, even with the state-of-the-art methods described above~\cite{salgan-360,baseline}. In this work, we show through experiments that this can be addressed by the data augmentation by rotating the ERP image in spherical coordinates added on top of the existing architecture~\cite{baseline}.
\section{Proposed method}
An overview of the proposed method is illustrated in~\Fref{teian_all}. 
Given a single \omni~image in ERP as input, the proposed method (1) predicts a saliency map using prior baseline networks~\cite{baseline} trained using our data augmentation technique by spherical random rotations, (2) extracts RoI candidates based on Selective Search~\cite{selective_search}, and (3) optimizes a set of RoIs based on our Salient-IoU evaluation score.

\subsection{Saliency map prediction using spherical random rotation augmentation}
\begin{figure}[t]
  \begin{center}
   \includegraphics[width = 80mm]{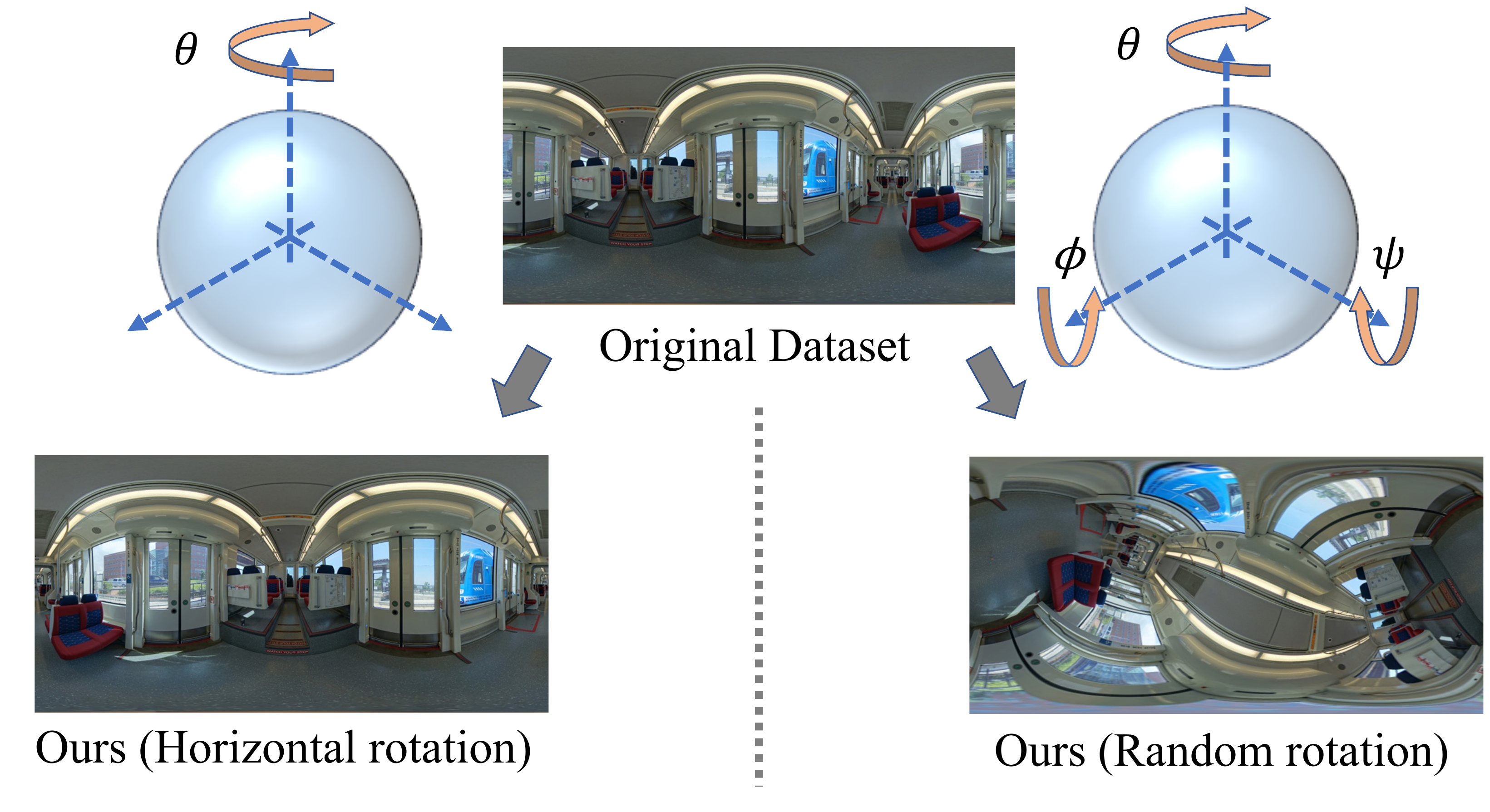}
  \end{center}
  \vspace{-2mm}
  \caption{Examples of random rotation data augmentation}
  \vspace{-2mm}
  \label{teian_rotate}
 \end{figure}

Based on its definition, the RoIs draw more attention from human viewers than other regions in the same image -- RoIs can be considered as the region of local maximum saliency. As introduced in~\Sref{sec:relatedwork_sal}, there are a few deep neural network models for this task but trained on mid-scale datasets~\cite{Rai2017,salient360-image} which contain less than two hundred pairs of a \omni~image and a corresponding gaze information where most salient regions are concentrated near the equator. N\"aively training a network on this training data inevitably results in this strong center bias and makes it difficult to extract RoIs apart from the equator.

To tackle both the problems of scarce of data and the strong center bias, we propose a simple but effective spherical data augmentation strategy. 
As illustrated in~\Fref{teian_rotate}, given a pair of \omni~images and gaze maps in ERP format, we first back-project ERP images onto the unit sphere, then apply random rotation, and project them back onto the ERP coordinates. 
In this random rotation, the image with its paired saliency map on sphere is randomly rotated around three axes individually: $\theta\in[-\pi,\pi]$ around the gravity direction, $\phi\in[-\frac{\pi}{2},\frac{\pi}{2}]$ around the grazing axis, and $\psi\in[-\pi,\pi]$ around the axis perpendicular to both rotation axes. 
After the rotation, the salient region around the equator moves away from the equator; 
therefore, the strong center bias is removed. 
In our experiments, we will show that simply applying the proposed method to the state-of-the-art single image \omni~saliency prediction model by Martin~\etal~\cite{baseline} dramatically improves the prediction accuracy even with mid-scale training examples.

\subsection{RoI extraction through Salient-IoU optimization}
The goal of our framework is to extract a predefined number of RoIs (\eg, $n=5$ in this study) as distinctively salient regions in the input \omni~image. 
However, most pixels in the predicted saliency map contain non-smooth, non-zero entries and simply thresholding the saliency values will result in the generation of a number of isolated regions.
Another possible strategy can be to first apply an object detection algorithm (\eg, \cite{RCNN}) to the input image and simply take top-$n$ objects with the highest saliency values (\ie, salient object detection). 
Nevertheless, this simple strategy cannot extract an RoI that includes multiple objects, and more importantly, the result is restricted to specific recognizable categories of objects. Instead, we take a two-step approach, which is composed of the extraction of candidate regions that are perceptually important and the optimization of selection of subset of RoIs that satisfy our criteria. 

\begin{figure}[t]
  \begin{center}
   \includegraphics[width = 80mm]{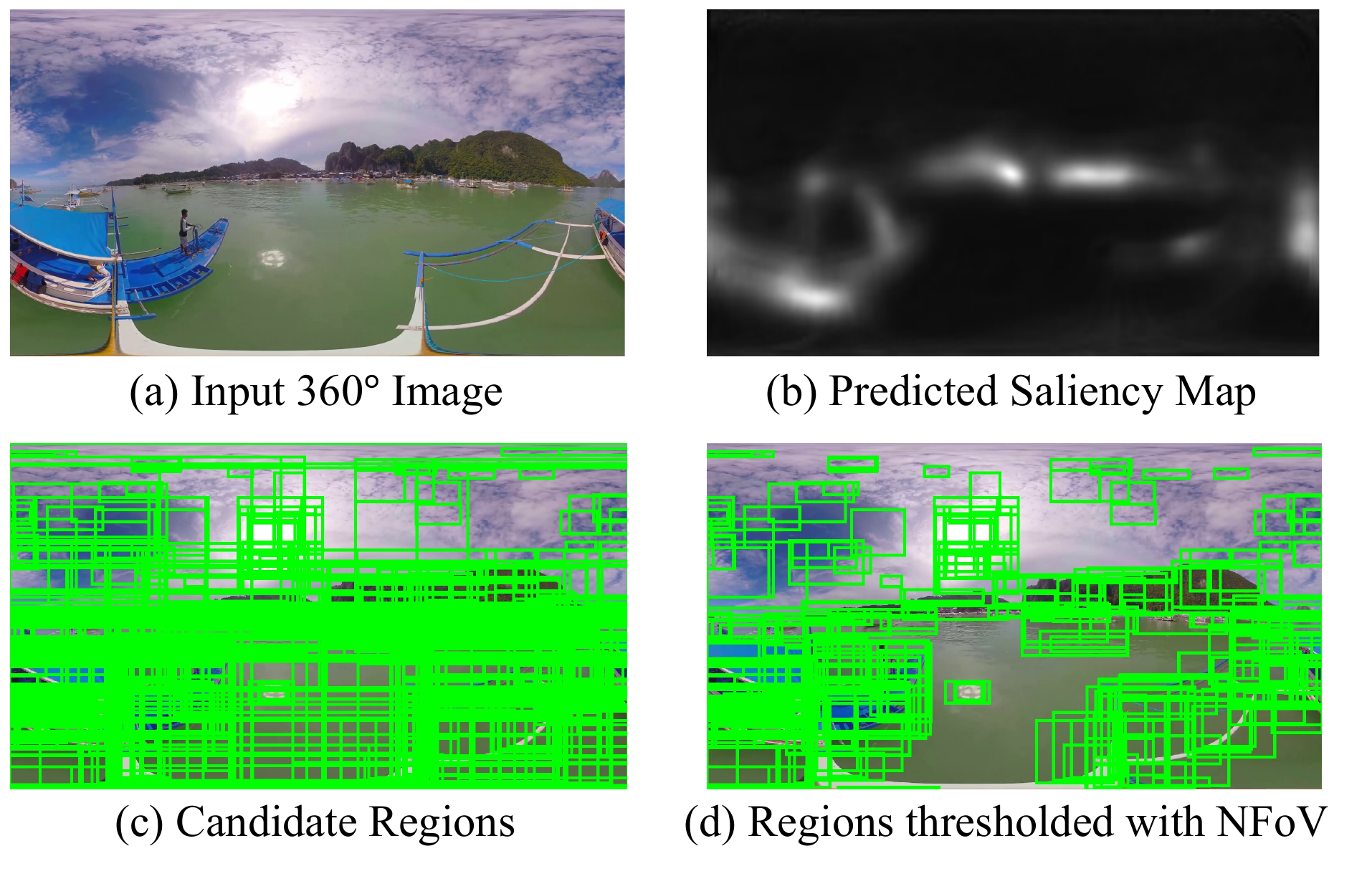}
  \end{center}
    \vspace{-4mm}
  \caption{Top row: Inputs for the RoI extraction. Bottom row:Before and after thresholding output regions of Selective Search in NFoV=65\textdegree}
  \vspace{-2mm}
  \label{roi_input}
 \end{figure}

In the first step, we apply Selective Search~\cite{selective_search} algorithm to the input \omni~image to extract the candidate regions. Selective Search, which is used for producing object proposals in the early time of object detection algorithms~\cite{RCNN}, greedily merges superpixels based on low-level features such as the color, texture, size, and fitness to extract perceptually coherent regions. 
Selective Search is a purely bottom-up approach that works for various scenes, unlike recent top-down region proposal neural networks~\cite{FasterRCNN} which prefer regions around specific recognizable categories of objects.  
The output of the Selective Search is illustrated in \Fref{roi_input}. 
The resultant region proposals are overlapping rectangles of various sizes distributed over an entire image. 
Because we assume that RoIs whose corresponding FoV is smaller than NFoV, we exclude regions where the latitude or longitude FoV is larger than NFoV (\ie, 65\textdegree) from the result of Selective Search. 

Using the predicted saliency map, our task is then to find the optimal subset of the candidate regions that maximally summarizes the entire \omni~image in a perceptually plausible manner. This is a non-trivial problem because if we only use top-$n$ regions of the average or summation of the saliency values, a number of overlapping regions will be extracted around the most salient area in the entire image. 
It is not necessarily desirable to extract regions only around the most salient area because the other important areas are most likely to be overlooked. Assume we want to extract $n$~RoI regions, it is desirable that they are different top-$n$ salient regions from different parts of the image. 
\begin{figure}[t]
\centering
    \includegraphics[width=0.78\linewidth]{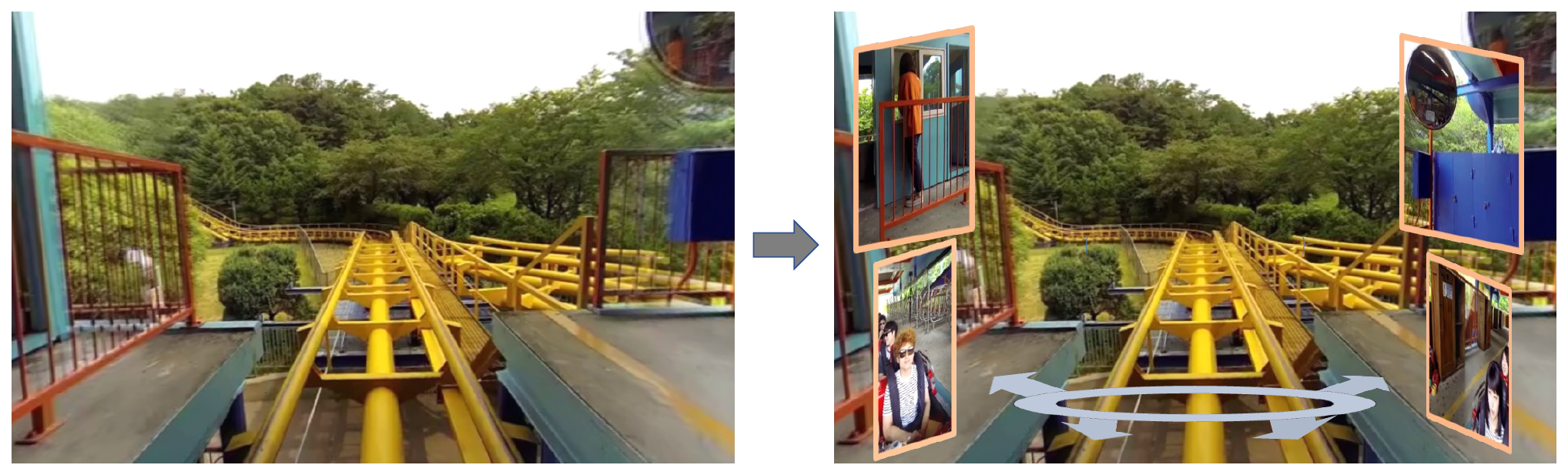}
    \caption{An example of \omni~ image viewer with the RoIs.}
    \vspace{-6mm}
    \label{gui}
\end{figure}
\begin{table*}[h]
\caption{Evaluation of saliency prediction with 
data augmentation}
\begin{center}
\scalebox{1.2}[1.2]{
\begin{tabular}{lcccccc}\hline
\multicolumn{1}{c}{}  & \multicolumn{1}{l}{AUC\_Judd↑} & \multicolumn{1}{l}{AUC\_Borji↑} & \multicolumn{1}{l}{NSS↑} & \multicolumn{1}{l}{CC↑} & \multicolumn{1}{l}{SIM↑} & \multicolumn{1}{l}{KLD↓} \\ \hline 
SalGAN360~\cite{salgan-360}              & 0.758                          & 0.703                           & \textbf{1.309}                    & 0.234                   & 0.466                    & 1.890                    \\

PC(Baseline)~\cite{baseline}              & 0.728                          & 0.693                           & 0.874                    & 0.432                   & 0.545                    & 1.074                    \\

Ours w/ random rotation      & \textbf{0.774}                          & \textbf{0.735}                           & 1.183                    & \textbf{0.584}                   & \textbf{0.609}                    & \textbf{0.842}  \\ \hline
Ours w/ horizontal rotation & 0.772                          & 0.730                           & 1.105                    & 0.553                   & 0.592                    & 0.949                    \\
\hline
\end{tabular}
}
\label{aug}
\end{center}
\end{table*}
\begin{figure*}[t]
\centering
\scalebox{1}[1]{
\begin{tabular}{ccccc|c}
\includegraphics[width=25mm]{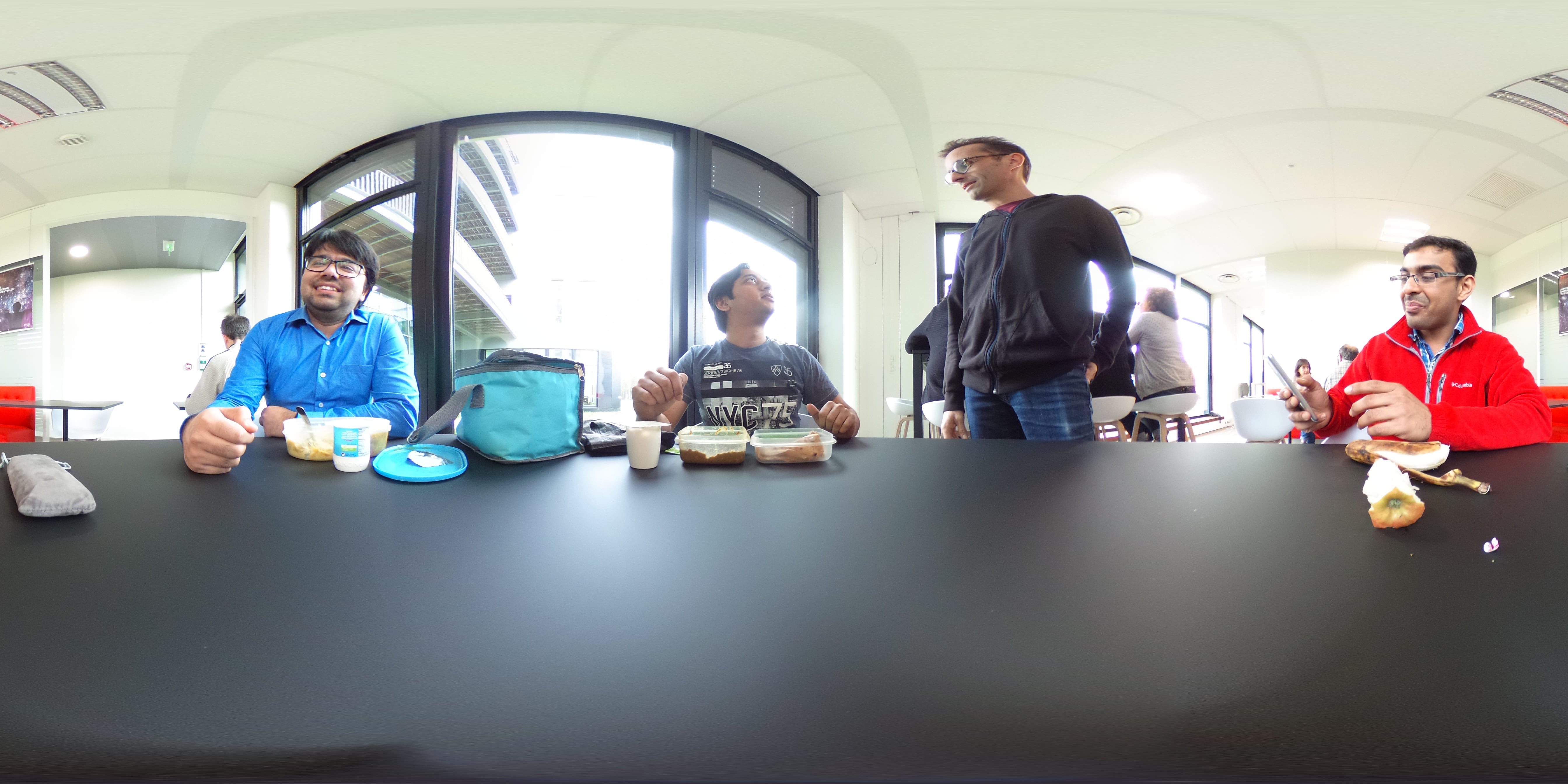}
&\includegraphics[width=25mm]{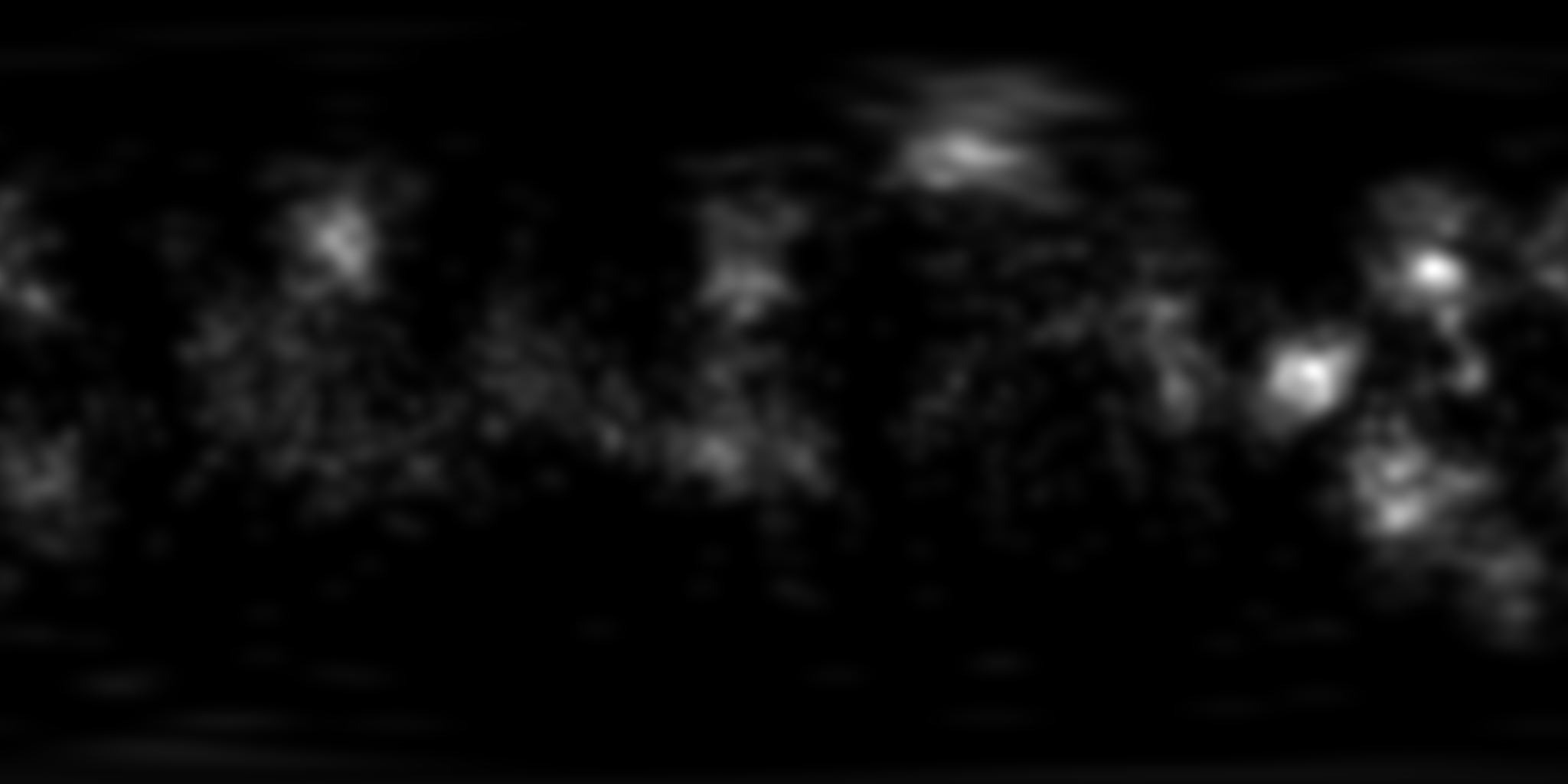}
&\includegraphics[width=25mm]{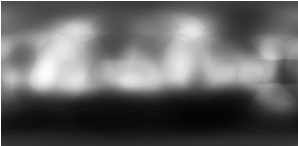}
&\includegraphics[width=25mm]{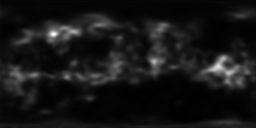}
&\includegraphics[width=25mm]{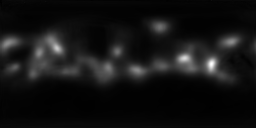}
&\includegraphics[width=25mm]{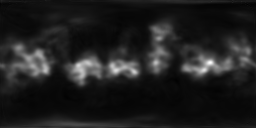}
\\

\footnotesize{Salient360!\cite{salient360-image}: P94}&\footnotesize{Ground Truth}&\footnotesize{SalGAN360\cite{salgan-360}}&\footnotesize{PC(Baseline)\cite{baseline}}&\footnotesize{Ours (Random rotation)}&\footnotesize{Ours (Horizontal rotation)}\\

\includegraphics[width=25mm]{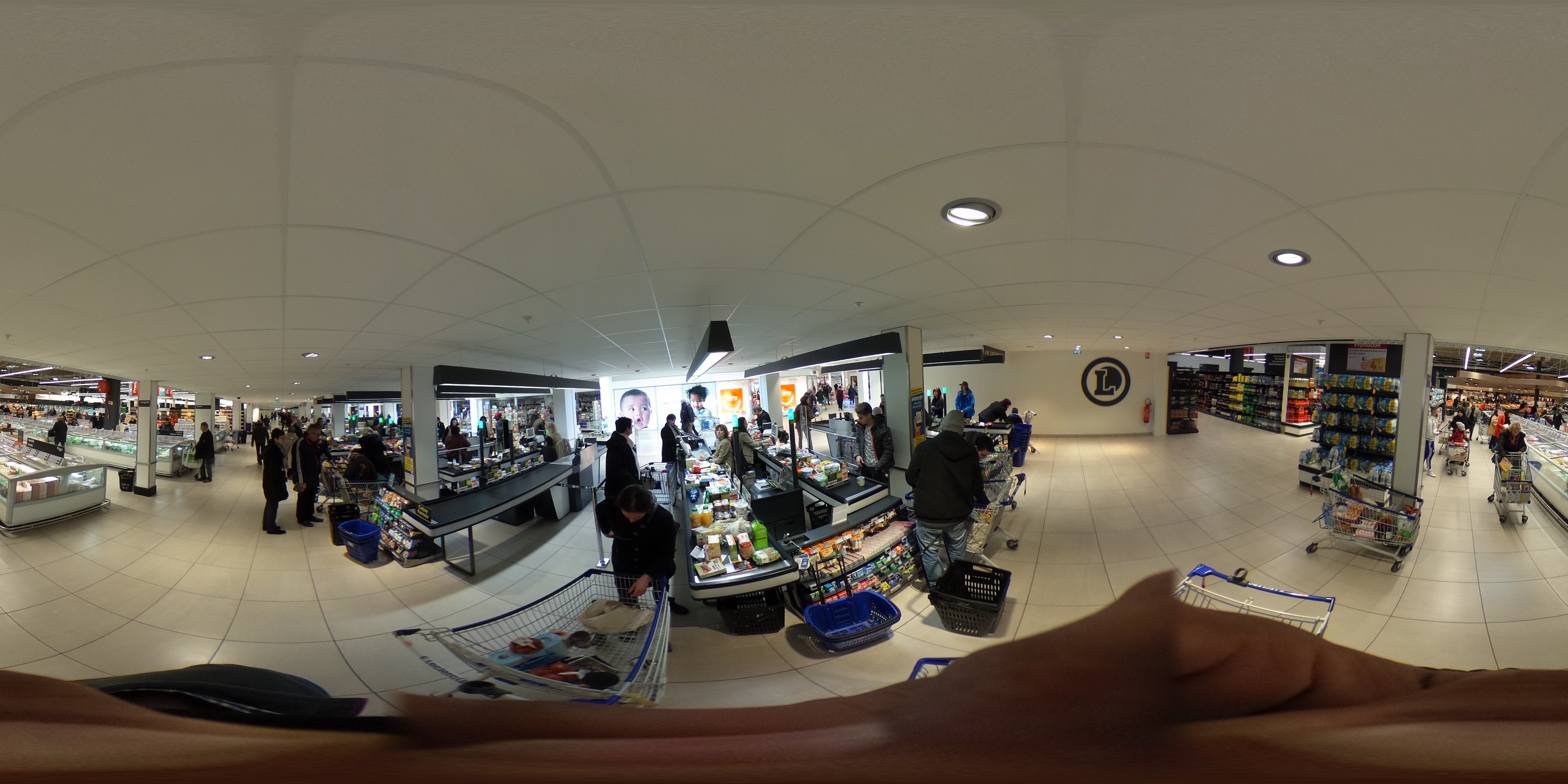}
&\includegraphics[width=25mm]{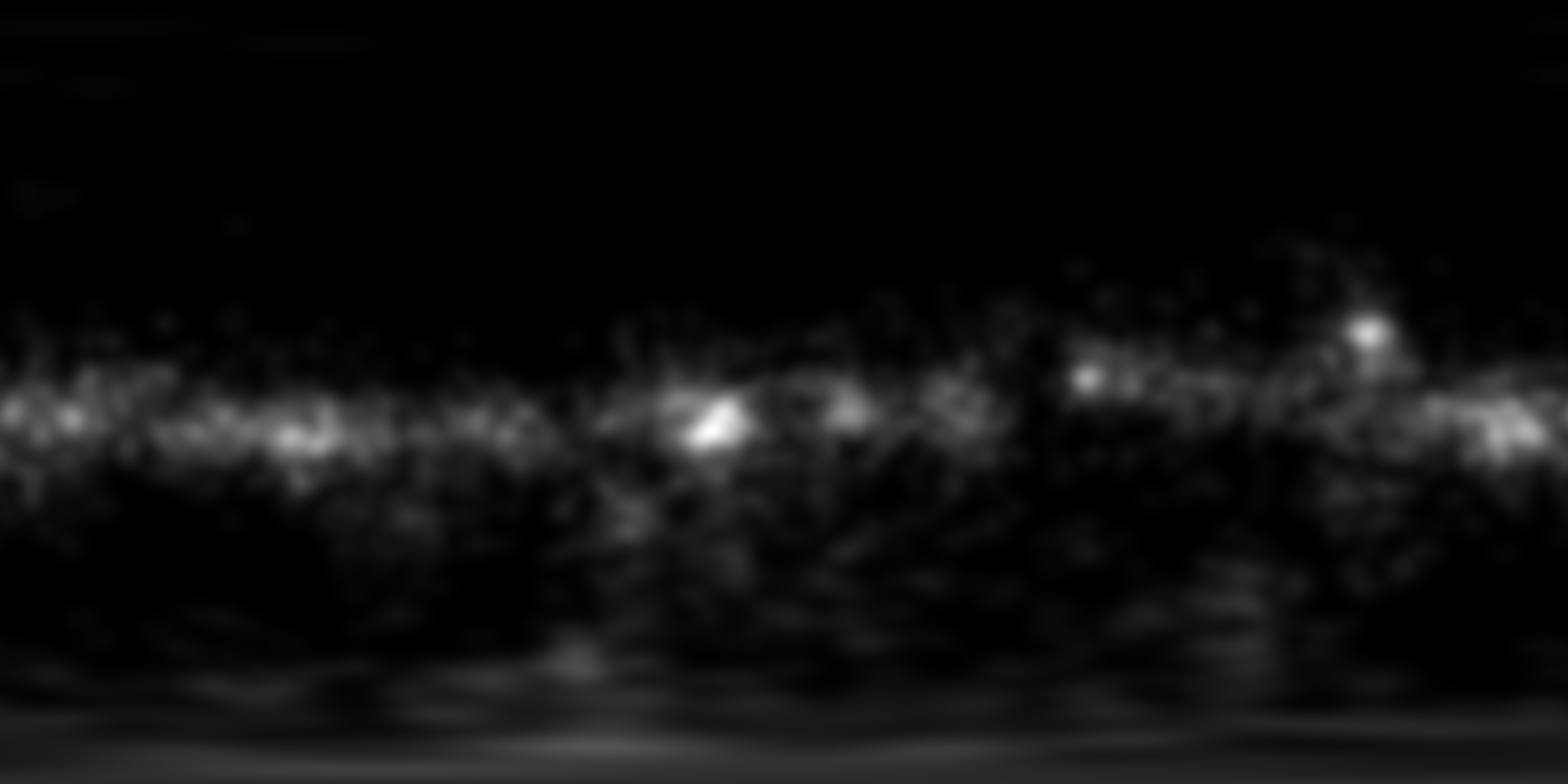}
&\includegraphics[width=25mm]{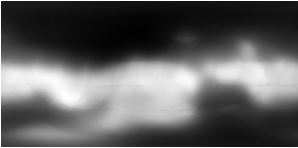}
&\includegraphics[width=25mm]{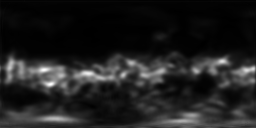}
&\includegraphics[width=25mm]{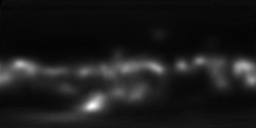}
&\includegraphics[width=25mm]{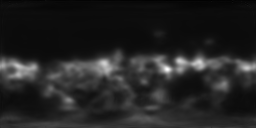}
\\
\footnotesize{Salient360!\cite{salient360-image}: P96}&\footnotesize{Ground Truth}&\footnotesize{SalGAN360\cite{salgan-360}}&\footnotesize{PC(Baseline)\cite{baseline}}&\footnotesize{Ours (Random rotation)}&\footnotesize{Ours (Horizontal rotation)}\\

\end{tabular}
}
\vspace{-2mm}
\caption{From left to right: input image of test data, ground truth of saliency map, results of training with SalGAN360~\cite{salgan-360}, results of training with the baseline~\cite{baseline}, results of training with data augmentation by random rotation of the proposed method, and results of training with data augmentation by horizontal rotation of the proposed method.}
\vspace{-2mm}
\label{result_saliency}
\end{figure*}

Then, we propose to extract of multiple saliency regions with a certain size that have less overlap with each other. 
Specifically, we minimize the cost function, namely Salient-IoU ($\gamma$) which evaluates a subset of $n$ regions  ($S$) from a set of all candidate regions ($R$) generated by Selective Search. SIoU considers both the sum of the saliency of the region and the overlap between the regions and gives a smaller value to a subset that is considered to be better as follows:
\begin{equation}
{\gamma}(S)=\frac{a}{n} \sum^{n}_{i=1}\frac{1}{g(I_{i})} +(1-a) \frac{1}{{}_n C _2} \sum_{i,j\in [1,n], i\neq j}{\mathrm{IoU}}(I_{i}, I_{j}),
\label{loss}
\end{equation}
where $I_{k\in[1,n]}\in S$ is the region included in $S$, and $g(I_{k})$ is the sum of the predicted saliency values in the region. 
Owing to the projection distortion of the ERP format, the local summation of saliency values is overestimated in high-latitude regions. 
Therefore, we multiply $w=\cos{\lambda}$ using the predicted saliency map, where $\lambda$ is the latitude $\lambda \in[-\frac{\pi}{2},\frac{\pi}{2}]$ used to maintain a constant pixel density on a surface of a unit sphere as presented in ~\cite{Huyen2018}.
The weighted saliency map is then $\ell_2$-normalized such that the summation of all saliency values become one. ${\rm IoU}(I_{i}, I_{j})$ is the Intersection-Over-Union (IoU)~\cite{RCNN}, which becomes one when two regions are completely overlapped and zero when they are not, and $a$ is the balancing weight ($a\in[0,1]$) which controls the  contribution of the saliency and the IoU. 
When $a$ is close to 1, regions of higher saliency are preferred, and when $a$ is close to zero, overlapping regions are less. 

The algorithm used to find the optimal set of RoIs based on Salient-IoU minimization is as follows:
Given a collection of regions (${R}$) from Selective Search which was filtered out by their FoV, the region subset ${S}$ is initialized with the top $n$ candidate regions that have the highest total saliency values.
and the elements in ${S}$ are removed from ${R}$. 
We then greedily replace one region in ${S}$ with another region in ${R}$, which has the highest total saliency values in ${R}$, one by one, and compare values of SIoU before and after the replacement. 
If the value of SIoU gets smaller after the replacement, ${S}$ and ${R}$ are updated -- the replacement is accepted and the element is removed from ${R}$. 
In the implementation, ${S}$ is an array which consists of $n$ regions.
With the element of index from zero to $n-1$ in ${S}$, replacement, calculation of SIoU values, and comparison are executed with the current target element which has the highest total saliency values in $R$.
If the all possible replacements between each element in ${S}$ and the current target element in ${R}$ are rejected, the current target element in ${R}$ is removed from ${R}$.
Next, the region with the highest total saliency value in ${R}$ becomes the current target.
This operation is repeated until there are no more candidates in ${R}$, and the final ${S}$ is the optimal RoI set. 
\subsection{\omni~ image viewer with extracted RoIs}
The extracted RoIs can be directly overlaid to the input \omni~image, however there should be more effective way to display summarized information to a viewer. For example, by applying the proposed method to each frame of a \omni~video, we can efficiently extract RoIs that changes in time series to the observer. While this is out of the main scope of this paper, we designed a mock-up GUI and displayed RoIs extracted from the video frames to the observer as shown in~\Fref{gui}. We empirically confirmed that the proposed GUI can efficiently teach viewers what important items and events exist outside viewer's field of view. The better GUI design based on our RoI extraction method is left for our future work.

\section{Experimental Results}
We conducted two main experiments to demonstrate the effectiveness of our proposed method. First, we evaluated our spherical random rotation augmentation for better training of the baseline saliency prediction network. Second, we conducted a user study to evaluate the cognitive appropriateness of obtained RoIs.

\subsection{Quantitative evaluation of saliency map prediction}
\noindent \textbf{Dataset details}: We used Salient360!~\cite{salient360,salient360-image} dataset for the saliency prediction network training and the evaluation. It contains 85 ERP images and corresponding ground truth saliency maps obtained using the eye tracker. We split entire pairs of \omni~image and saliency map into 78 for training and seven (\ie, P91, P93-P98) for the test. 

\noindent \textbf{Evaluation metrics}:
We used six evaluation metrics listed below -- five of them were used in the Grand Challenge of ICME'17 and ICME'18 of Salient360!\cite{salient360} and AUC\_Borji~\cite{auc_borji} was included as well.
\begin{itemize}
    \item Normalized Scanpath Saliency (NSS)
    \item Pearson's Correlation Coefficient (CC)
    \item Similarity or histogram intersection (SIM)
    \item Kullback-Leibler divergence (KLD)
   \item Area under ROC Curve by Judd (AUC\_Judd)
   \item Area under ROC Curve by Borji (AUC\_Borji)
\end{itemize}

Note that we followed the evaluation framework in salient360!~\cite{salient360}. Specifically, predicted saliency maps were weighted by their latitude to avoid overestimating the error in the high-latitude region because the sampling point at high-latitudes on the unit sphere are stretched horizontally on the ERP image.

\noindent \textbf{Implementation Details}:
We evaluated our data augmentation method by the random spherical rotation using one of the state-of-the-art \omni~saliency prediction network by Martin~\etal~ (Baseline, PanoramicConv, PC)\cite{baseline} as the backbone network without changing their original implementation. In addition to our data augmentation technique, Martin~\etal~used seven augmentation methods (\ie, three flips and additive Gaussian noise, Poisson noise, salt-pepper noise and speckle noise). It is a current state of the art, and we used them in the evaluation as well.

These seven augmentations cannot alleviate the problem of the strong center bias of \omni~ images.
, which is a special for saliency prediction of \omni~ images.
To remove the effects of the strong center bias, we used our random rotation augmentation in addition to the seven augmentations.
During training, the input \omni~image in ERP format was downsampled to 256$\times$128 and the parameters were optimized with the Momentum SGD~\cite{sgd_momentum} and Spherical Mean Squared Error (MSE) Loss~\cite{saliency-video}.
The hyperparameters for the training are as follows:
epoch=10000, batch size=32, learning rate=$10^{-4}$,
momentum=0.9, and weight decay=$10^{-5}$.
The entire evaluation framework was implemented using PyTorch~\cite{pytorch} where the network was trained and tested on a single NVIDIA Quadro RTX 8000 with 48GB of memory.

\noindent \textbf{Results}: The results are shown in Table \ref{aug}. The prediction by Martin~\etal~without our spherical rotation augmentation is shown as the baseline. Our method was implemented by using it as a backbone and additionally applied random spherical rotation augmentation (Proposal w/ random). For the ablation study, we also showed the result in which the training data was rotated only around the gravity axis (Proposal w/ horizontal). Note that the rotation was only applied to the training data, but not to the test data. 
Because  Martin~\etal~'s and Chao~\etal~\cite{salgan-360}'s models' codes are publicly available among the the state of the art methods, we used Chao~\etal~'s model for the comparison with the same setting of Martin~\etal's. Papers\cite{salgcn, zhu2019prediction} also used Chao~\etal~'s model as a comparison method.

For all evaluation metrics, the random rotation augmentation was shown to most effectively improve the performance, and its accuracy is significantly better than that of the baseline. We also see that the horizontal rotation is not enough.   

Figure \ref{result_saliency} shows a visualization of the obtained saliency.
It can be confirmed that the baseline method tends to predict high saliency values near the equator owing to the strong center bias of the training data, whereas the proposed method can predict the saliency robustly even within a high latitude region.
Besides, even if our horizontal rotation is applied to the dataset, predicted saliency values tend to be blurred and spread within the middle latitude region because of the strong center bias.
In contrast, when we use the random rotation, the blurred area becomes denser, and it shows the random rotation is necessary for a better prediction.

\subsection{Evaluation of single \omni~image multiple RoI prediction}
\begin{figure}[t]
  \begin{center}
   \includegraphics[width = 80mm]{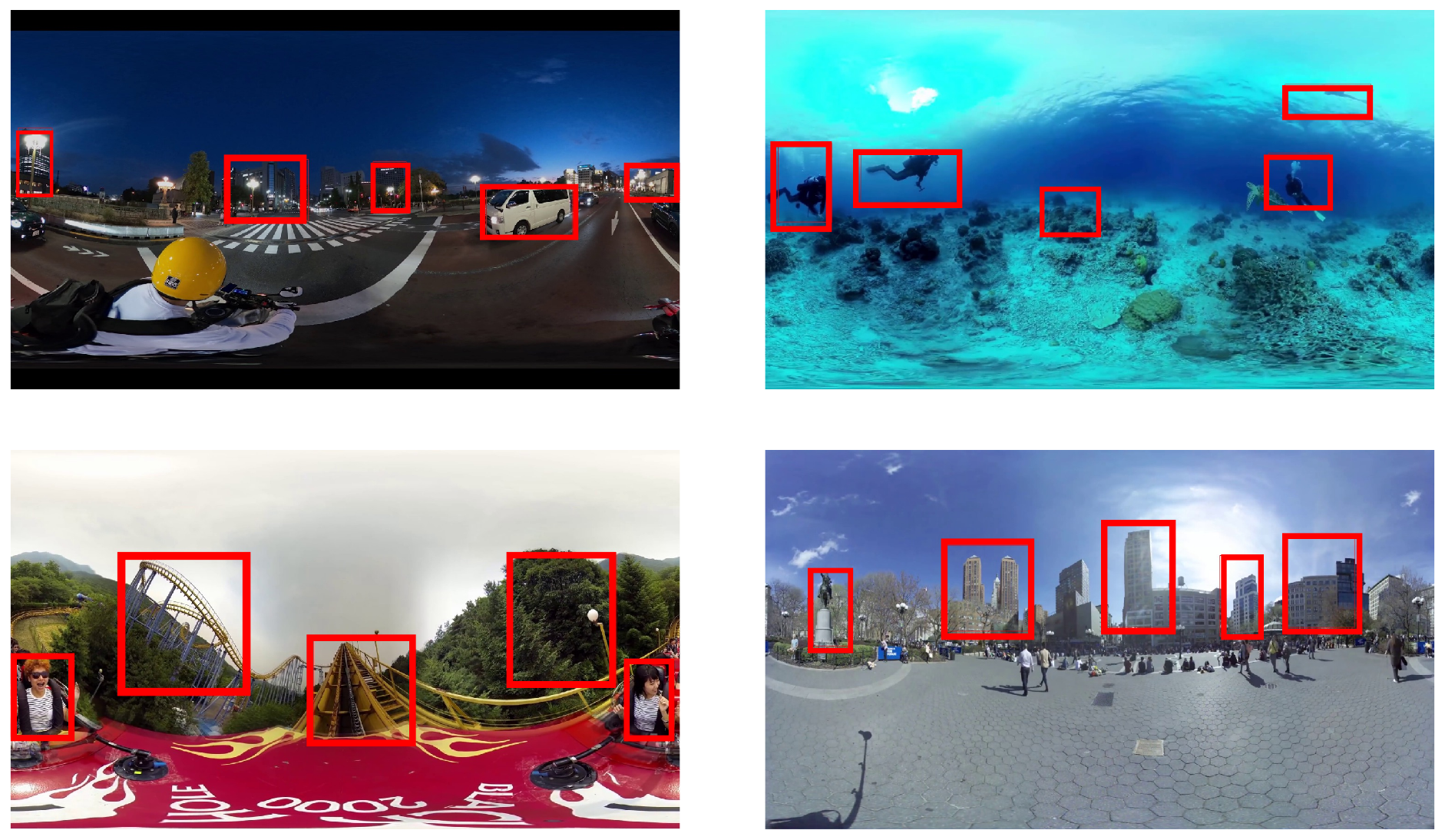}
  \end{center}
  \caption{Examples of user-annotated RoIs dataset for evaluation.}
  \vspace{-5mm}
  \label{dataset_example}
 \end{figure}
\noindent \textbf{Dataset construction}: There is no existing dataset for evaluating multiple RoI prediction task from a single~\omni~image, therefore we constructed the evaluation dataset which consists of pairs of \omni~image and RoIs annotated by human. Some examples in our dataset are shown in \Fref{dataset_example}. 

Our dataset consists of forty five ~\omni~images extracted from five \omni~outdoor video clips on YouTube~\cite{Youtube} and corresponding RoI annotation by three different persons. To annotate RoIs on each~\omni~image, we recruited crowd workers using Amazon Mechanical Turk~\cite{amt} to ask to annotate five impressive regions in each \omni~ image of the dataset. The dataset construction process is detailed as follow.

First, a cloud worker was instructed to look around an omnidirectional image on the web browser through cropped perspective view by freely changing the view direction and remember the scene. Actual instructions are as follows: (1) Look around the image by dragging the cursor. (2) Remember the parts of the image that are particularly impressive. (3) Take at least 30 seconds. Make sure not missing the ceiling or floor. 

Next, we showed the same~\omni~image but in ERP format to the worker and asked him/her to draw five bounding boxes around impressive parts. The instructions were as follows: 
(1) Use the bounding box tool to draw boxes around the area that are impressive to you. 
Draw a rectangle using your mouse over the area.
(2) Each rectangle should be less than 1/4 of the total image.
(3) This is not an object detection annotation, so you can enclose areas where there are no objects.
It is okay if the areas overlap each other.

A total of 14 crowd workers participated in the annotation.
We did not impose any restrictions on the qualifications of the crowd workers, but only approved those workers who spent more than 60 seconds on the task.
Our dataset includes 50 images in total, and each image was annotated by 3 persons, resulting in 150 pairs of \omni~image and user-annotated RoIs.
Excluding the inappropriate 5 images such as a blackout scene, we finally constructed an evaluation dataset of 45 images and 135 RoI annotations. Since RoIs in an image is highly subjective in nature, we didn't merge three annotations of the same~\omni~image in evaluation.

\begin{figure*}[t]
\centering
    \includegraphics[width=0.78\linewidth]{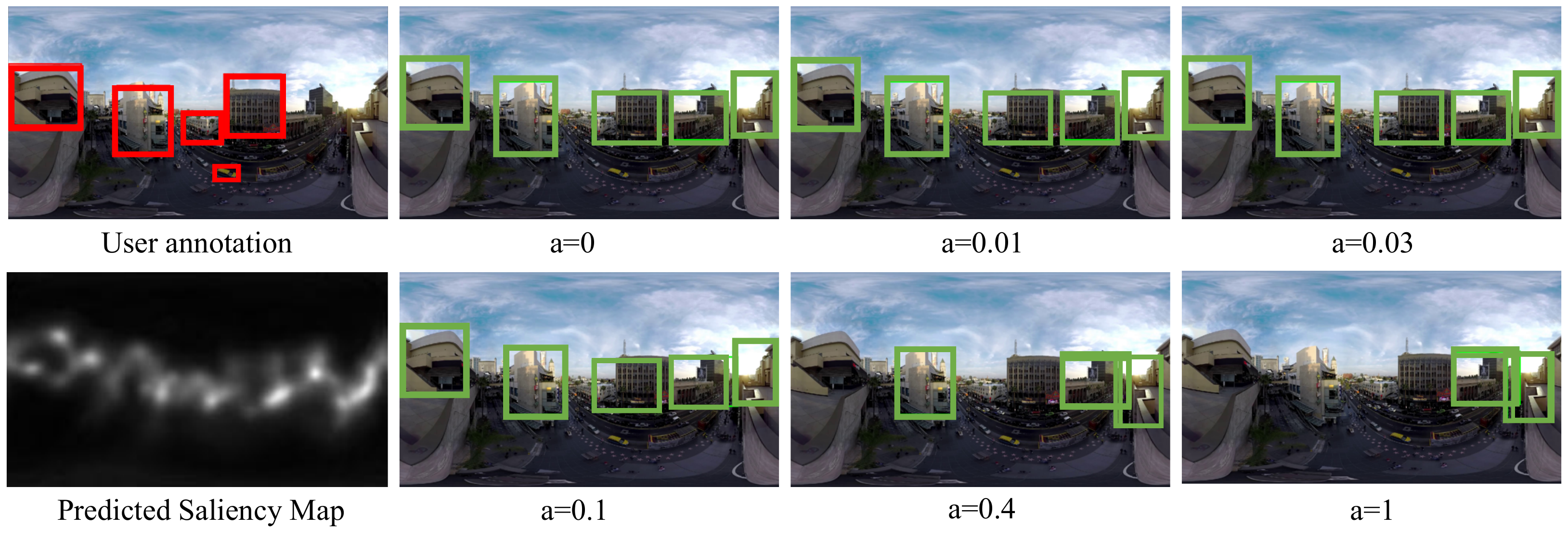} 
    \includegraphics[width=0.78\linewidth]{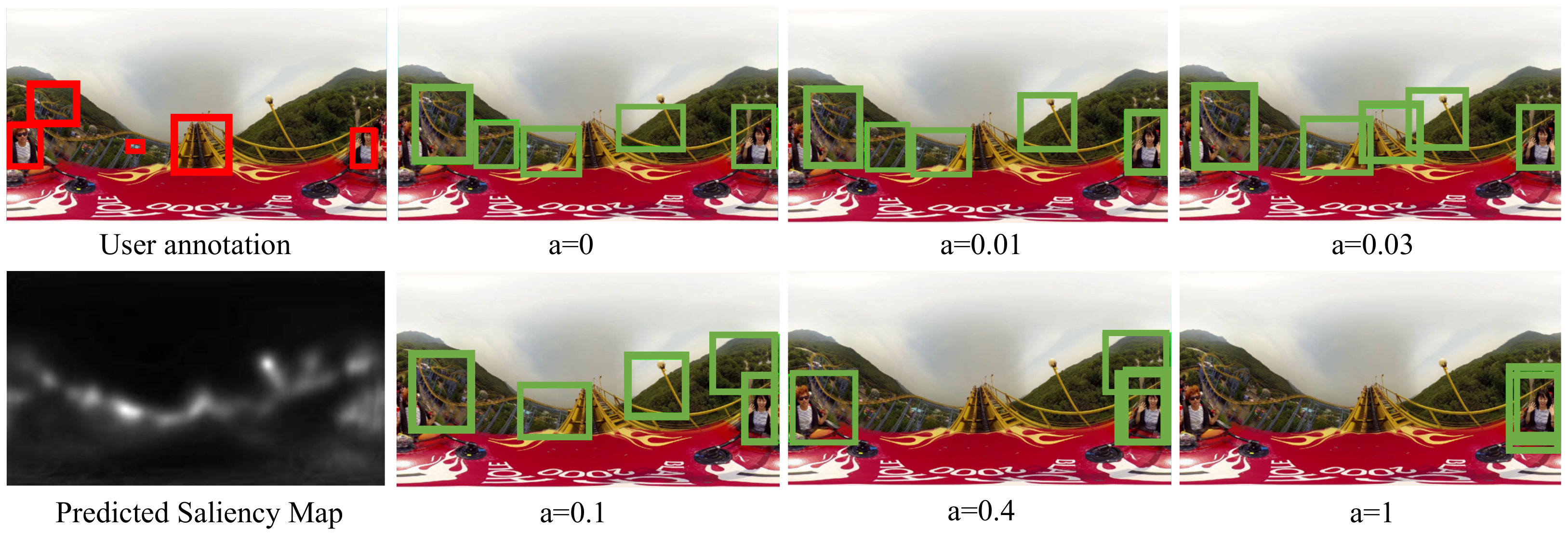} 
    \includegraphics[width=0.78\linewidth]{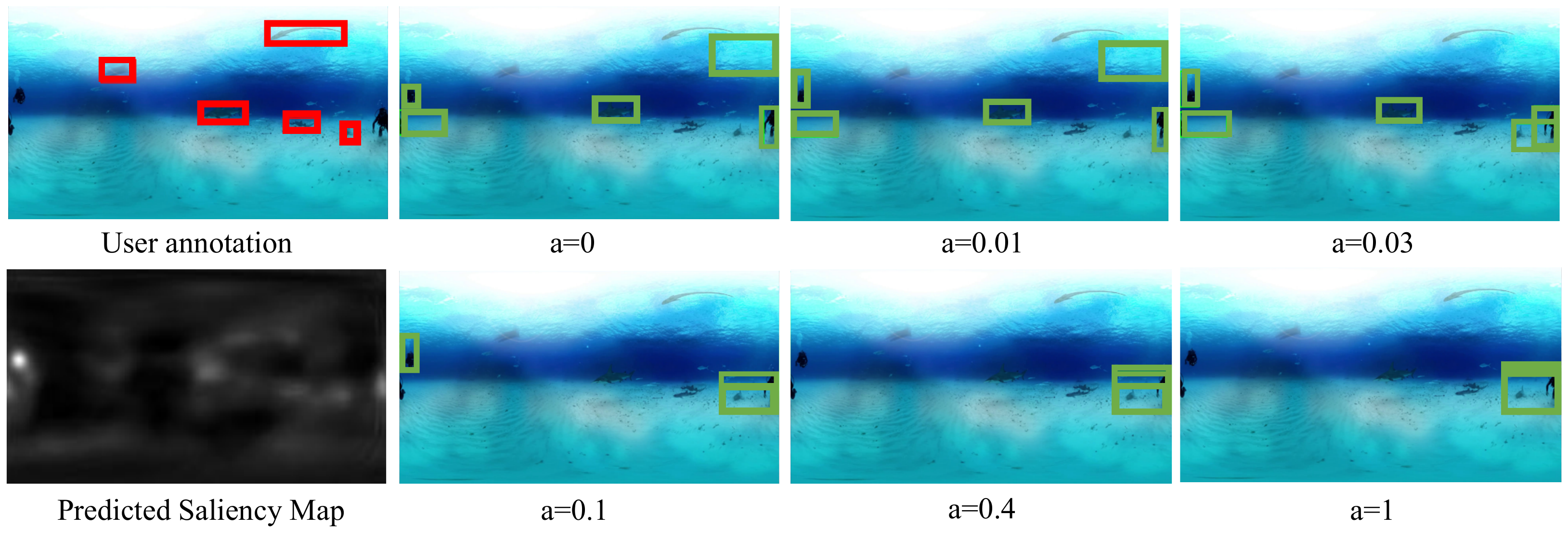}
    \caption{Examples of RoIs obtained by our framework with different $a$s.}
    \label{results_per_a}
\end{figure*}

\begin{figure}[t]
\centering
\scalebox{0.9}[0.8]{
\includegraphics[width=\linewidth]{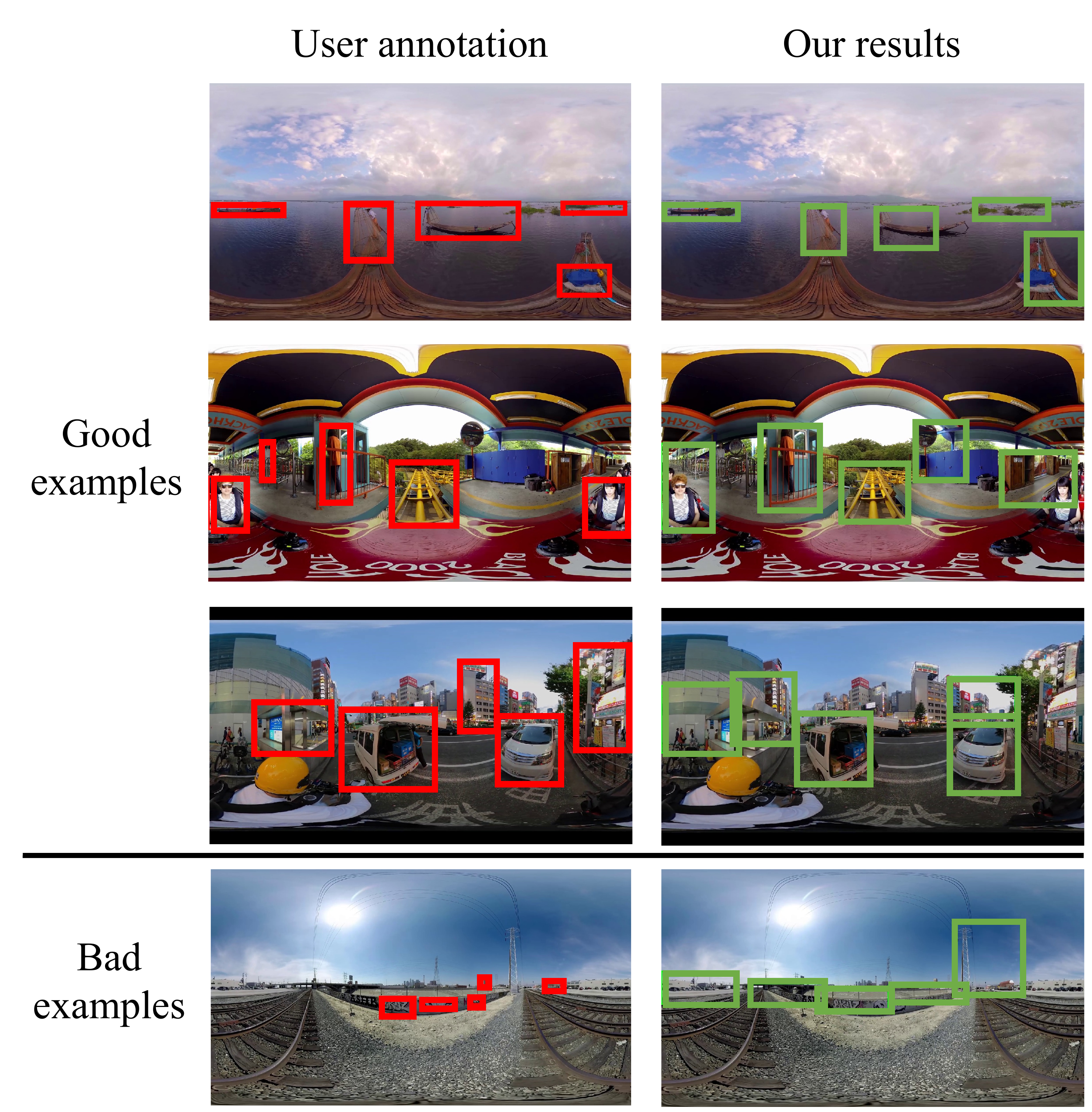}
\vspace{0mm}
}
\caption{Comparison of user annotated RoIs and our detected RoIs}
\label{result_roi}
\end{figure}
\begin{figure}[t]
    \centering
    \includegraphics[width=\linewidth]{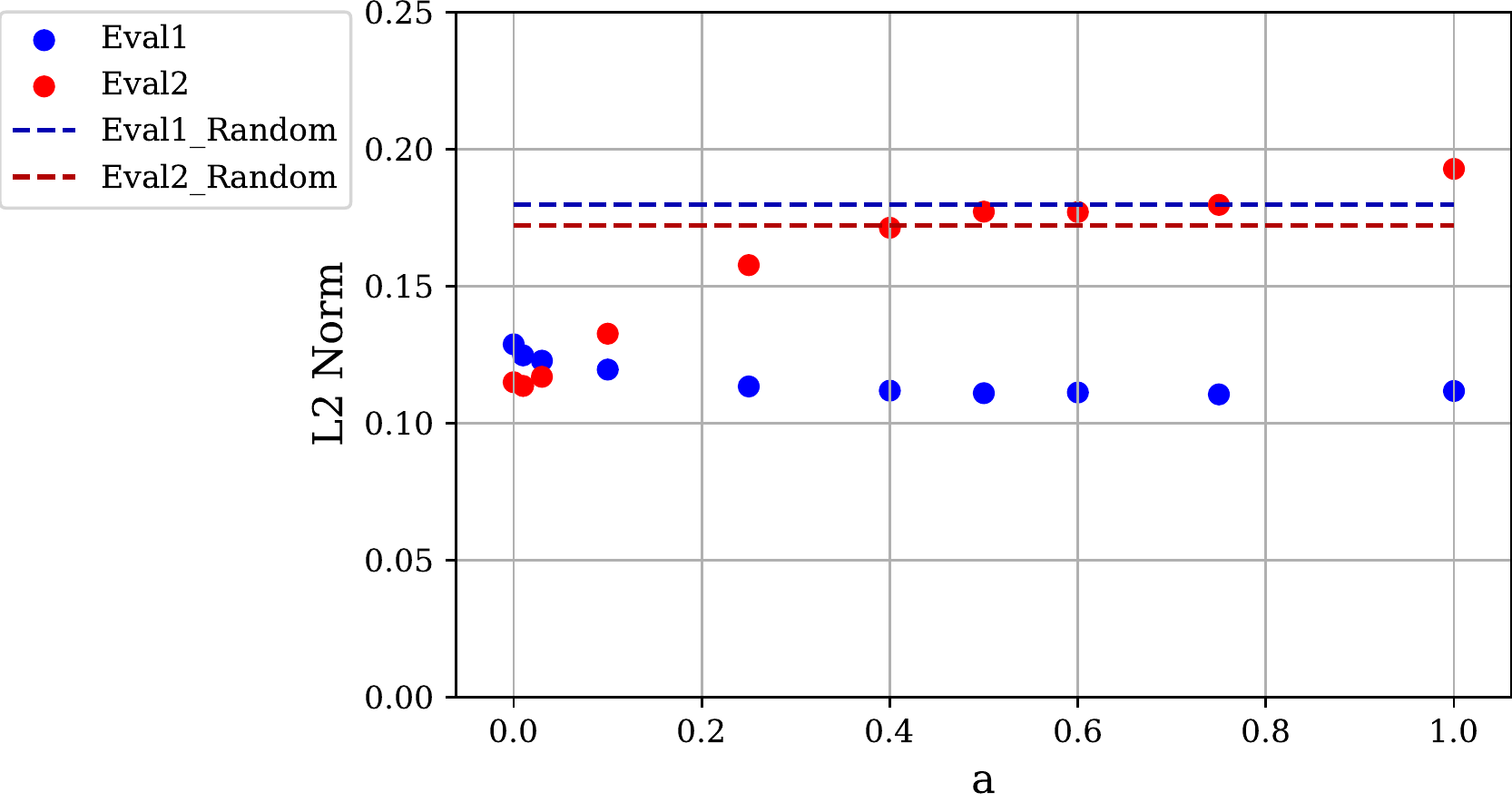}\\
    (a)\\
    \centering
    \includegraphics[width=\linewidth]{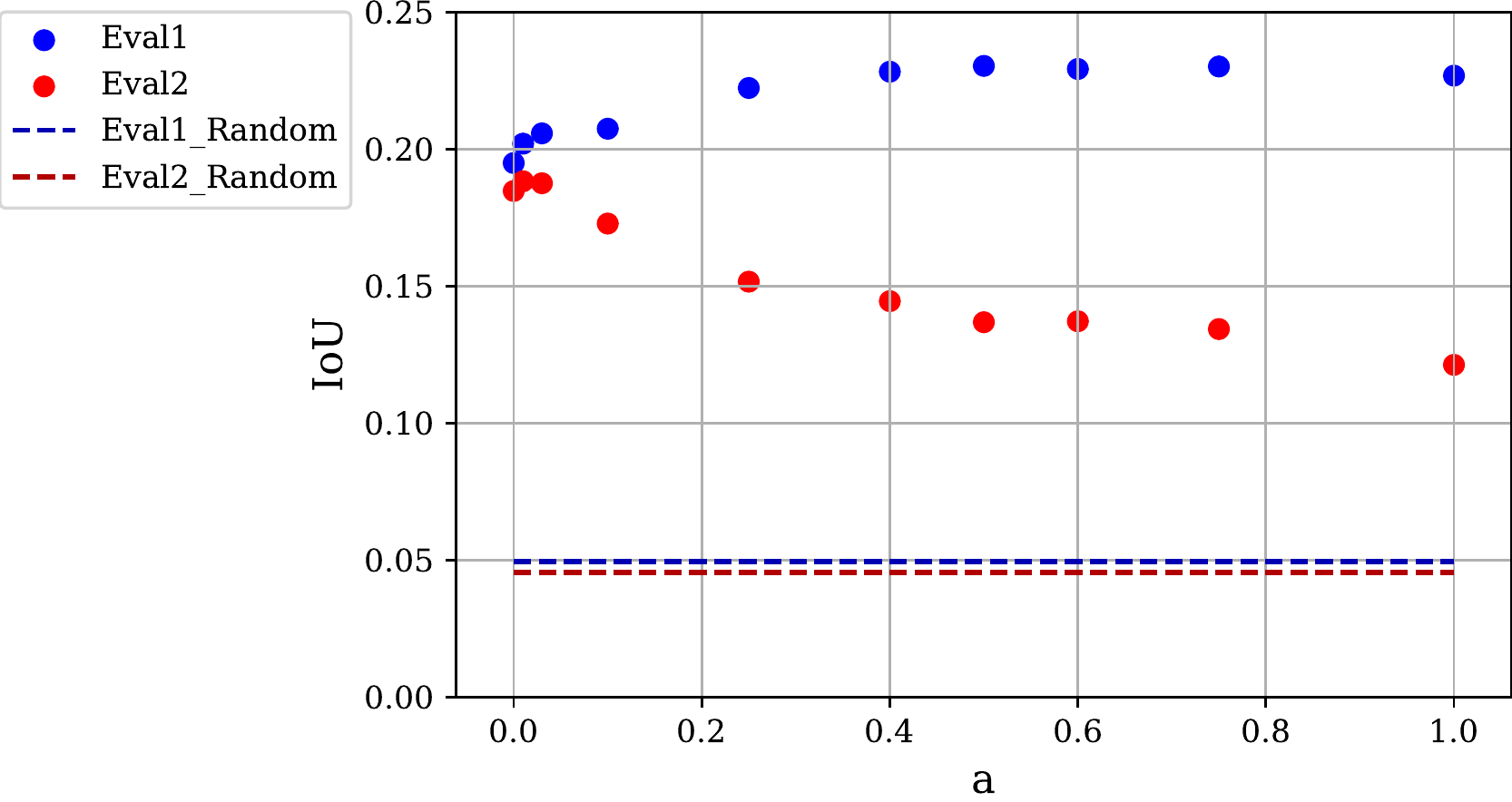}\\
    (b)
  \caption{Comprisons between the the user annotated RoIs and our detected RoIs for varying $a$s. (a) L2 norm, (b) IoU}
    \label{eval_iou}
    \vspace{-2mm}
\end{figure}

\noindent \textbf{Qualitative evaluation}:
Some examples of our result applied to the constructed dataset are shown in~\Fref{results_per_a}. The predicted RoIs (green color boxes) are overlaid on the input~\omni~image with different choices of $a$ in~\Eref{loss}. We observed that the prediction results were reasonably close to one of the user-annotated RoIs. As expected, only considering the saliency information (\ie, $a=1$) gave significantly overlapped RoIs, and we couldn't detect multiple RoIs. On the other hand, spatially distributed RoIs were extracted when we consider both visual saliency and IoU. In failure cases, as shown in the bottom of Figure \ref{result_roi}, images with few objects tended to produce RoIs that were significantly different from the user annotation. This is probably due to the fact that the training dataset for the saliency prediction does not include very small salient items.

\noindent \textbf{Quantitative evaluation}
We conducted a quantitative comparison between user-annotated RoIs ($S_{User}$) and the RoIs predicted by the proposed method ($S_{Pred}$).
The evaluation metrics for the quantitative comparison were the normalized Euclidean distance (L2 norm) on the image between the center points of the predicted and user-annotated RoIs, and the IoU between them.
Considering that the right and left edges of the ERP image are connected, 
the longest distance is $D=\sqrt{\frac{W}{2}^2 + H^2}$ when the number of vertical pixels in the ERP image is $H$ and the number of horizontal pixels is $W$.
The normalized Euclidean distance $D$ between the regions $\alpha=(x_\alpha,y_\alpha,w_\alpha,h_\alpha)$ and the region $\beta=(x_\beta,y_\beta,w_\beta,h_\beta)$ is computed as follows:
\begin{equation}
    D = \frac{\sqrt{{\rm min}(|x_\alpha-x_\beta|,W-|x_\alpha-x_\beta|)^2+(y_\alpha - y_\beta)^2}}{\sqrt{\frac{W}{2}^2 + H^2}}.
\end{equation}

For the IoU, because Selective Search\cite{selective_search} is originally an algorithm for perspective images, it does not take into account a candidate region that crosses the right to the left edge of an ERP image.

We considered the following two methods for paring human-annotated regions and predicted regions for comparison. ${S_{User}}$ and ${S_{Pred}}$ means a set of five RoIs by human-annotation and our prediction, respectively.
\begin{itemize}
    \item For each RoI in ${S_{Pred}}$, choose the region of ${S_{User}}$ that gives the best evaluation, and average all of the best evaluations. (Eval1)
    \item For each RoI in ${S_{User}}$, choose the region of ${S_{Pred}}$ that gives the best evaluation, and average all of the best evaluations. (Eval2)
\end{itemize}
Eval1 and Eval2 are similar to precision and recall, respectively.
The graphs of the results for these evaluations are shown in \Fref{eval_iou}. 
In addition to the human-annotation and our prediction, we show a random selection in \Fref{eval_iou}. The random selection randomly chose five regions from  the regions obtained by Selective Search. Its evaluation is shown by dashed lines in the figures.

It is noteworthy that when the saliency was more weighted in Salient-IoU, L2 norm slightly decreased in Eval1 and increased in Eval2. 
This indicates that the RoIs predicted by emphasizing only saliency and neglecting the degree of overlap between RoIs were concentrated in specific high salient region, and the RoIs did not well distribute within the images.
In other words, although the prediction with $a=1$ accurately select the highest salient region, but "overlook" is not avoided. 
These results are consistent with the observations in the qualitative evaluation presented in the previous section.

\subsection{User study for evaluating predicted RoIs}
We conducted subjective evaluation to verify the quality of predicted RoIs. For presenting predicted/annotated RoIs on the NFoV perspective image, we converted RoIs defined on the ERP coordinates to ones on perspective images without projection distortions. In this conversion, we set a tangent plane contacting on the unit sphere at the center coordinate of the RoI on the ERP image, and project points on the unit sphere using the RoI as the projection plane.
When the image size of the projection plane is set to $H_p$ and $W_p$, the following equations are obtained for $H_p$ and $W_p$:
\begin{equation}
H_p = 2r\tan~(\frac{B_h}{H}\frac{\pi}{2})
\end{equation}
\begin{equation}
W_p = 2r\tan~(\frac{B_w}{W}\pi)
\end{equation}
where $r$ is the radius of the unit sphere, $H$ and $W$ are the height and width in pixels of the ERP image, respectively. 
$B_h\in(0,H)$ and $B_w\in(0,\frac{W}{2})$ are the height and width of each RoI, respectively. We extracted five perspective projection images from a \omni~ image based on the resulting RoIs. 
This resulted in 135 pairs of a ERP image and a set of five RoIs for evaluation. 

For the user study, we recruited the subjects on Amazon Mechanical Turk\cite{amt} to view each \omni~ image in a browser for 30s, and answer a questionnaire.
The protocol is that after showing the \omni~ images to the cloud workers for 30s, we showed them three groups of images: 
\begin{enumerate}
\item[A] Five perspective RoI images of our user-annotated RoIs dataset.
\item[B] Five perspective RoI images predicted by the proposed method.
\item[C] Five randomly chosen regions from candidates obtained by Selective Search.
\end{enumerate}
We created the following question: 
"Which of the three sets of regions do you think is the most impressive in the \omni~ image?"
We also provided a choice "tie."
Workers were not told which group each set of images corresponded to, and the order of the options was changed each time.
We received three responses for each of the 135 sets and repeated these experiments six times with different Sailent-IoU parameters $a$ while maintaining $n=5$.
We only approved workers who spent at least 60 seconds per one image.
We did not limit the maximum number of tasks that one worker could participate in, and thus the number of workers differed from 42 to 67 as the value of $a$ changed.
In each \omni~image, we got answers from different three workers.

The selection rates for each image group are shown in Table \ref{result_amt}. 
Interestingly, the RoI predicted by the proposed method was rated higher than the user-annotated RoI, that is, manual choice. 
According to the results of $\chi^2$ tests, on the condition of $a=0, 0.01, 0.03, 0.1, 0.4$, there is statistical significance between the frequencies of User annotation and Our results.
In addition, the RoI detected with weight on both saliency and IoU was more preferred, and it shows SIoU is the reasonable evaluation function for RoI detection.

\begin{table}[t]
\caption{Selection rate for each image group}
\label{user_study}
\centering
\scalebox{0.9}[0.9]{
\begin{tabular}{lcccc}\hline
      & Random & User annotation & Our results    & Tie     \\ \hline
a=0   & 58/405                   &  119/405**   & \textbf{218}/405**  & 10/405                    \\
a=0.01 &  48/405                   & 124/405**   & \textbf{215}/405**  & 18/405                    \\
a=0.03 & 43/405                   & 110/405**    & \textbf{236}/405** &  16/405                   \\
a=0.1 & 51/405                   & 115/405**    & \textbf{229}/405**  & 10/405                    \\
a=0.4 & 48/405                   & 157/405*\,\,    & \textbf{191}/405*\,\,   & 9/405                    \\
a=1   & 68/405                  & 159/405\,\,\,  & \textbf{164}/405\,\,\, & 14/405    \\
\hline
*:p<0.1,**:p<0.01
\end{tabular}
}
\label{result_amt}
\vspace{-4mm}
\end{table} 
\section{conclusion}
In this study, we tackle the problem of predicting the regions of interest (RoI) from a single \omni~ image as a set of perspective projection images with variable FoV and free positioning relationships. 
We proposed an algorithm to predict the optimal RoI set based on the saliency map predicted from \omni~ images and an evaluation function considering both the saliency value and the IoU in candidate regions obtained through Selective Search.
To train the network to predict the saliency map, we proposed the random rotation data augmentation to 
overcome the strong center bias of the training data, 
and showed a significant improvement in performance over the baseline~\cite{baseline}. 
We created RoI dataset, and evaluated the predected RoIs in quantitatively, qualitatively with it.
In user study, we show that our algorithm can predict reasonable RoIs.



\bibliographystyle{ieeetr}
\bibliography{references}

\begin{thebibliography}{10}

\bibitem{pano2vid}
Y.-C. Su, D.~Jayaraman, and K.~Grauman, ``Pano2vid: Automatic cinematography
  for watching 360\textdegree~ videos,'' in {\em Proceedings of the Asian
  Conference on Computer Vision (ACCV)}, 2016.

\bibitem{snap_angle}
B.~Xiong and K.~Grauman, ``Snap angle prediction for 360° panoramas,'' in {\em
  Proceedings of the European Conference on Computer Vision (ECCV)}, September
  2018.

\bibitem{selective_search}
J.~R. Uijlings, K.~E. Van De~Sande, T.~Gevers, and A.~W. Smeulders, ``Selective
  search for object recognition,'' {\em International journal of computer
  vision}, vol.~104, no.~2, pp.~154--171, 2013.

\bibitem{saliency_neuro}
R.~Desimone and J.~Duncan, ``Neural mechanisms of selective visual attention,''
  {\em Annual review of neuroscience}, vol.~18, no.~1, pp.~193--222, 1995.

\bibitem{itti}
L.~{Itti}, C.~{Koch}, and E.~{Niebur}, ``A model of saliency-based visual
  attention for rapid scene analysis,'' {\em IEEE Transactions on Pattern
  Analysis and Machine Intelligence}, vol.~20, no.~11, pp.~1254--1259, 1998.

\bibitem{deepgaze2}
M.~Kummerer, T.~S.~A. Wallis, L.~A. Gatys, and M.~Bethge, ``Understanding low-
  and high-level contributions to fixation prediction,'' in {\em Proceedings of
  the IEEE International Conference on Computer Vision (ICCV)}, Oct 2017.

\bibitem{salgan-360}
F.-Y. Chao, L.~Zhang, W.~Hamidouche, and O.~Deforges, ``Salgan360: Visual
  saliency prediction on 360 degree images with generative adversarial
  networks,'' in {\em 2018 IEEE International Conference on Multimedia \& Expo
  Workshops (ICMEW)}, pp.~01--04, IEEE, 2018.

\bibitem{baseline}
D.~Martin, A.~Serrano, and B.~Masia, ``Panoramic convolutions for
  360\textdegree~ single-image saliency prediction,'' in {\em CVPR Workshop on
  Computer Vision for Augmented and Virtual Reality}, 2020.

\bibitem{salnet360}
R.~Monroy, S.~Lutz, T.~Chalasani, and A.~Smolic, ``Salnet360: Saliency maps for
  omni-directional images with cnn,'' {\em Signal Processing: Image
  Communication}, vol.~69, pp.~26--34, 2018.
\newblock Salient360: Visual attention modeling for 360° Images.

\bibitem{salgcn}
H.~Lv, Q.~Yang, C.~Li, W.~Dai, J.~Zou, and H.~Xiong, ``Salgcn: Saliency
  prediction for 360-degree images based on spherical graph convolutional
  networks,'' in {\em Proceedings of the 28th ACM International Conference on
  Multimedia}, MM '20, (New York, NY, USA), p.~682–690, Association for
  Computing Machinery, 2020.

\bibitem{salient360}
J.~Gutiérrez, E.~David, A.~Coutrot, M.~P.~D. Silva, and P.~L. Callet,
  ``Introducing un salient360! benchmark: A platform for evaluating visual
  attention models for 360 contents,'' in {\em International Conference on
  Quality of Multimedia Experience (QoMEX)}, 2018.

\bibitem{salient360-image}
Y.~Rai, J.~Guti{\'e}rrez, and P.~Le~Callet, ``A dataset of head and eye
  movements for 360 degree images,'' in {\em Proceedings of the 8th ACM on
  Multimedia Systems Conference}, pp.~205--210, 2017.

\bibitem{amt}
``{Amazon Mechanical Turk}.'' https://www.mturk.com/, 2022.

\bibitem{Su2017}
Y.~{Su} and K.~{Grauman}, ``Making 360° video watchable in 2d: Learning
  videography for click free viewing,'' in {\em 2017 IEEE Conference on
  Computer Vision and Pattern Recognition (CVPR)}, pp.~1368--1376, 2017.

\bibitem{deep_360_pilot}
H.~{Hu}, Y.~{Lin}, M.~{Liu}, H.~{Cheng}, Y.~{Chang}, and M.~{Sun}, ``Deep 360
  pilot: Learning a deep agent for piloting through 360° sports videos,'' in
  {\em 2017 IEEE Conference on Computer Vision and Pattern Recognition (CVPR)},
  pp.~1396--1405, 2017.

\bibitem{Lee2018}
S.~{Lee}, J.~{Sung}, Y.~{Yu}, and G.~{Kim}, ``A memory network approach for
  story-based temporal summarization of 360° videos,'' in {\em 2018 IEEE/CVF
  Conference on Computer Vision and Pattern Recognition (CVPR)},
  pp.~1410--1419, 2018.

\bibitem{Wang2020}
M.~Wang, Y.-J. Li, W.-X. Zhang, C.~Richardt, and S.-M. Hu,
  ``{Transitioning360}: Content-aware {NFoV} virtual camera paths for 360°
  video playback,'' in {\em International Symposium on Mixed and Augmented
  Reality (ISMAR)}, 2020.

\bibitem{Bur2006}
A.~{Bur}, A.~{Tapus}, N.~{Ouerhani}, R.~{Siegwart}, and H.~{Hugli}, ``Robot
  navigation by panoramic vision and attention guided fetaures,'' in {\em 18th
  International Conference on Pattern Recognition (ICPR)}, vol.~1,
  pp.~695--698, 2006.

\bibitem{Rai2017}
Y.~Rai, J.~Guti\'{e}rrez, and P.~Le~Callet, ``A dataset of head and eye
  movements for 360 degree images,'' in {\em Proceedings of the 8th ACM on
  Multimedia Systems Conference}, MMSys'17, (New York, NY, USA), p.~205–210,
  Association for Computing Machinery, 2017.

\bibitem{Erwan2018}
E.~David, J.~Gutiérrez, A.~Coutrot, M.~Perreira Da~Silva, and P.~Le~Callet,
  ``A dataset of head and eye movements for 360° videos,'' pp.~432--437, 06
  2018.

\bibitem{Cheng2018}
H.~{Cheng}, C.~{Chao}, J.~{Dong}, H.~{Wen}, T.~{Liu}, and M.~{Sun}, ``Cube
  padding for weakly-supervised saliency prediction in 360° videos,'' in {\em
  2018 IEEE/CVF Conference on Computer Vision and Pattern Recognition (CVPR)},
  pp.~1420--1429, 2018.

\bibitem{Zhang2018}
Z.~Zhang, Y.~Xu, J.~Yu, and S.~Gao, ``Saliency detection in 360° videos,'' in
  {\em The European Conference on Computer Vision (ECCV)}, September 2018.

\bibitem{salgan}
J.~Pan, C.~C. Ferrer, K.~McGuinness, N.~E. O'Connor, J.~Torres, E.~Sayrol, and
  X.~G. i~Nieto, ``Salgan: Visual saliency prediction with generative
  adversarial networks,'' 2018.

\bibitem{RCNN}
R.~{Girshick}, J.~{Donahue}, T.~{Darrell}, and J.~{Malik}, ``Rich feature
  hierarchies for accurate object detection and semantic segmentation,'' in
  {\em 2014 IEEE Conference on Computer Vision and Pattern Recognition (CVPR)},
  pp.~580--587, 2014.

\bibitem{FasterRCNN}
S.~{Ren}, K.~{He}, R.~{Girshick}, and J.~{Sun}, ``Faster r-cnn: Towards
  real-time object detection with region proposal networks,'' {\em IEEE
  Transactions on Pattern Analysis and Machine Intelligence}, vol.~39, no.~6,
  pp.~1137--1149, 2017.

\bibitem{Huyen2018}
H.~T.~T. TRAN, C.~T. PHAM, N.~P. NGOC, A.~T. PHAM, and T.~C. THANG, ``A study
  on quality metrics for 360 video communications,'' {\em IEICE Transactions on
  Information and Systems}, vol.~E101.D, no.~1, pp.~28--36, 2018.

\bibitem{auc_borji}
A.~{Borji}, D.~N. {Sihite}, and L.~{Itti}, ``Quantitative analysis of
  human-model agreement in visual saliency modeling: A comparative study,''
  {\em IEEE Transactions on Image Processing}, vol.~22, no.~1, pp.~55--69,
  2013.

\bibitem{sgd_momentum}
N.~Qian, ``On the momentum term in gradient descent learning algorithms,'' {\em
  Neural Networks}, vol.~12, no.~1, pp.~145--151, 1999.

\bibitem{saliency-video}
Z.~Zhang, Y.~Xu, J.~Yu, and S.~Gao, ``Saliency detection in 360° videos,'' in
  {\em Proceedings of the European Conference on Computer Vision (ECCV)},
  September 2018.

\bibitem{pytorch}
A.~Paszke, S.~Gross, F.~Massa, A.~Lerer, J.~Bradbury, G.~Chanan, T.~Killeen,
  Z.~Lin, N.~Gimelshein, L.~Antiga, A.~Desmaison, A.~Kopf, E.~Yang, Z.~DeVito,
  M.~Raison, A.~Tejani, S.~Chilamkurthy, B.~Steiner, L.~Fang, J.~Bai, and
  S.~Chintala, ``Pytorch: An imperative style, high-performance deep learning
  library,'' in {\em Advances in Neural Information Processing Systems 32}
  (H.~Wallach, H.~Larochelle, A.~Beygelzimer, F.~d\textquotesingle
  Alch\'{e}-Buc, E.~Fox, and R.~Garnett, eds.), pp.~8024--8035, Curran
  Associates, Inc., 2019.

\bibitem{zhu2019prediction}
Y.~Zhu, G.~Zhai, X.~Min, and J.~Zhou, ``The prediction of saliency map for head
  and eye movements in 360 degree images,'' {\em IEEE Transactions on
  Multimedia}, vol.~22, no.~9, pp.~2331--2344, 2019.

\bibitem{Youtube}
``{YouTube}.'' https://www.youtube.com/, 2022.

\end{thebibliography}

\begin{IEEEbiography}[{\includegraphics[width=1in,height=1.25in,clip,keepaspectratio]{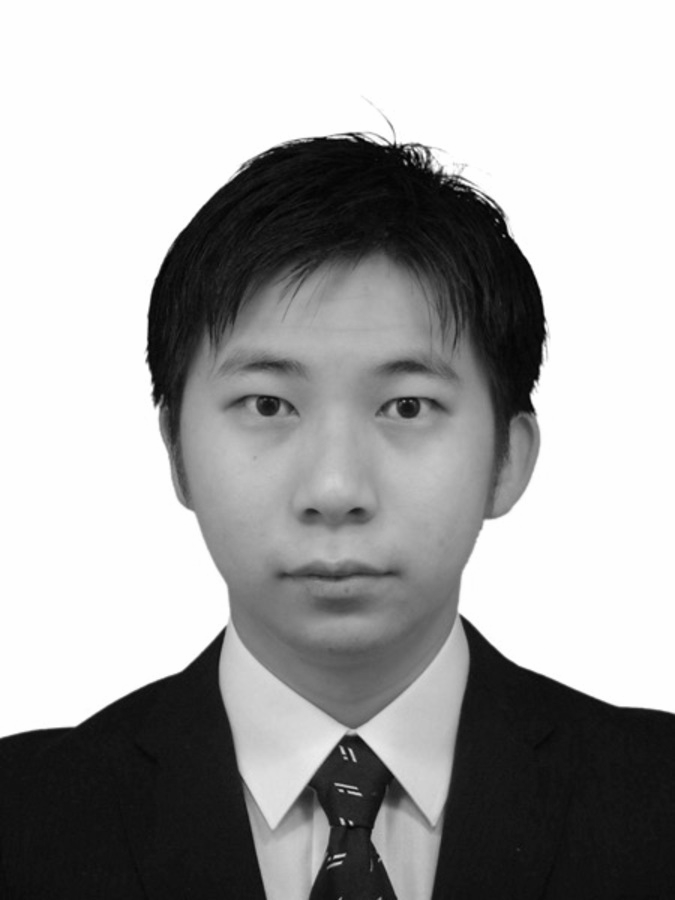}}]{Yuuki Sawabe} (Student M'20) 
received the B.S. degree in electrical and electronic engineering from University of Tokyo, Tokyo, Japan in 2021. 
He is currently pursuing the M.S. degree in information science and technology at Graduate School of Information Science and Technology, The University of Tokyo, Tokyo, Japan.
He is mainly interested in computer vision, \omni~ image processing, and VR/AR. 
\end{IEEEbiography}

\begin{IEEEbiography}[{\includegraphics[width=1in,height=1.25in,clip,keepaspectratio]{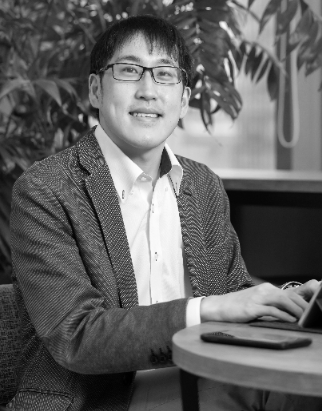}}]{Satoshi Ikehata} (M'11) received the B.A. of Psychology in 2009, and the M.S. and Ph.D of Information Studies in 2011 and 2014 from the University of Tokyo. He worked as a post-doc researcher in Washington University in St. Louis from 2014 to 2016. Currently, he is an assistant professor at National Institute of Informatics. His main interests lie in 3-D computer vision; physics-based 3-D reconstruction, VR/AR, indoor/outdoor scene understanding and human 3-D cognition and perception.
\end{IEEEbiography}

\begin{IEEEbiography}[{\includegraphics[width=1in,height=1.25in,clip,keepaspectratio]{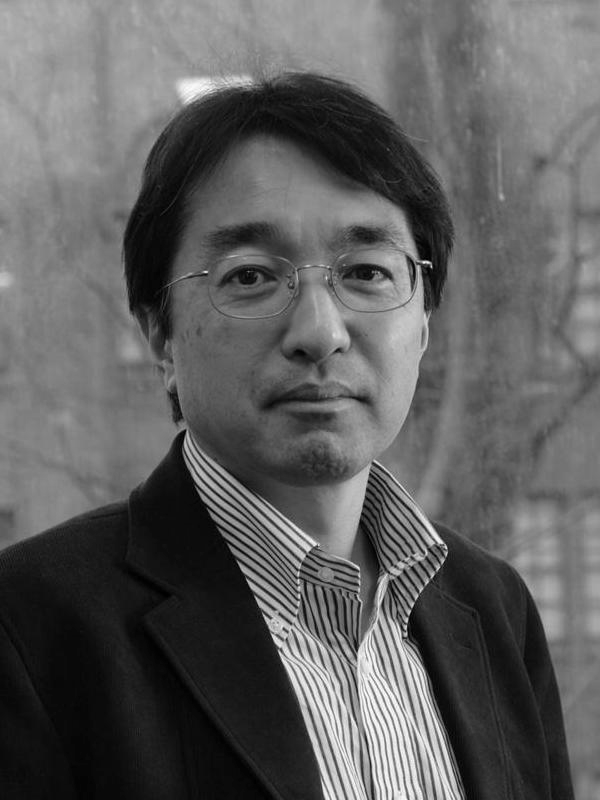}}]{Kiyoharu Aizawa} (Student M'83-M'88-SM'15-F'16) received the B.E., the M.E., and the Dr.Eng. degrees in Electrical Engineering all from the University of Tokyo, in 1983, 1985, 1988, respectively. He is currently a Professor at Department of Information and Communication Engineering of the University of Tokyo. He was a Visiting Assistant Professor at University of Illinois from 1990 to 1992. His research interest is in multimedia applications, image processing and computer vision.
He received the 1987 Young Engineer Award and the 1990, 1998 Best Paper Awards, the 1991 Achievement Award, 1999 Electronics Society Award from IEICE Japan, and the 1998 Fujio Frontier Award, the 2002 and 2009 Best Paper Award, and 2013, 2020 Achievement award from ITE Japan. He received the IBM Japan Science Prize in 2002.
He is on Editorial Boards of IEEE MultiMedia, ACM TOMM.
He served as the Editor in Chief of Journal of ITE Japan, an Associate Editor of IEEE Trans. Image Processing, IEEE Trans. CSVT and IEEE Trans. Multimedia. He was a president of ITE and ISS society of IEICE, 2019 and 2018, respectively. He has served a number of international and domestic conferences; he was a General co-Chair of ACM Multimedia 2012 and ACM ICMR2018. He is a Fellow of IEEE, IEICE, ITE and a council member of Science Council of Japan.
\end{IEEEbiography}

\EOD

\end{document}